\documentclass[sigconf, nonacm]{acmart}

\usepackage{makecell}
\usepackage{setspace}
\usepackage{makecell}
\usepackage{multirow}
\usepackage{amsmath}
\usepackage{graphicx}
\usepackage{subfigure}
\usepackage[algoruled]{algorithm2e}

\newcolumntype{L}{>{\raggedright\arraybackslash}X}
\newcolumntype{C}{>{\centering\arraybackslash}X}
\newcolumntype{P}[1]{>{\centering\arraybackslash}p{#1}}

\AtBeginDocument{%
  \providecommand\BibTeX{{%
    \normalfont B\kern-0.5em{\scshape i\kern-0.25em b}\kern-0.8em\TeX}}}

\setcopyright{acmcopyright}
\copyrightyear{2021}
\acmYear{2021}

\acmConference[DEEM '21]{DEEM '18: Workshop on Data Management for End-to-End Machine Learning}{June 20, 2021}{Virtual}

\begin{document}

\title{Understanding and Optimizing Packed Neural Network Training for Hyper-Parameter Tuning}

\author{Rui Liu, Sanjay Krishnan, Aaron J. Elmore, Michael J. Franklin}
\email{{ruiliu, skr, aelmore, mjfranklin}@uchicago.edu}
\affiliation{%
  \institution{University of Chicago}
  \city{Chicago}
  \state{IL}
  \country{USA}
}


\begin{abstract}
As neural networks are increasingly employed in machine learning practice, how to efficiently share limited training resources among a diverse set of model training tasks becomes a crucial issue. To achieve better utilization of the shared resources, we explore the idea of jointly training multiple neural network models on a single GPU in this paper. We realize this idea by proposing a primitive, called \texttt{pack}. 
We further present a comprehensive empirical study of \texttt{pack} and end-to-end experiments that suggest significant improvements for hyperparameter tuning.
%
%
The results suggest: (1) packing two models can bring up to $40\%$ performance improvement over unpacked setups for a single training step and the improvement increases when packing more models; (2) the benefit of the \texttt{pack} primitive largely depends on a number of factors including memory capacity, chip architecture, neural network structure, and batch size; (3) there exists a trade-off between packing and unpacking when training multiple neural network models on limited resources; (4) a \texttt{pack}-aware Hyperband is up to 2.7$\times$ faster than the original Hyperband, with this improvement growing as memory size increases and subsequently the density of models packed. 
\end{abstract}



\maketitle

\section{Introduction}
\label{intro}
\sloppy
The successes of AI are in part due to the adoption of neural network models which can place immense demand on computing infrastructure. It is increasingly the case that a diverse set of model training tasks share limited training resources. The long-running nature of these tasks and the large variation in their size and complexity make efficient resource sharing a crucial concern.
The concerns are compounded by an extensive trial-and-error development process where parameters are tuned and architectures have tweaked that result in a large number of trial models to train.
Beyond the monetary and resource costs, there are long-term questions of economic and environmental sustainability~\citep{strubell2019energy, schwartz2019green}.

Efficiently sharing the same infrastructure among multiple training tasks, or multi-tenant training, is proposed to address the issue~\citep{xiao2018gandiva, krishnan2019artificial, narayanan2019pipedream, mahajan2019themis}. The role of a multi-tenancy framework is to stipulate policies and constraints on how contended resources are partitioned and tasks are placed on physical hardware.
Most existing approaches divide resources at the granularity of full devices (e.g., an entire GPU)~\citep{DBLP:conf/usenix/JeonVPQXY19}. 
Such a policy can result in low resource utilization due to its coarse granularity. For example, models may greatly vary in size, where the largest computer vision models require multiple GBs of GPU memory~\citep{canziani2016analysis} but mobile-optimized networks use a significantly smaller space~\citep{DBLP:conf/cvpr/SandlerHZZC18}. Given that GPUs today have significantly more on-board memory than in the past (e.g., up to 32 GB in commercial offerings), if a training workload consists of a large number of small neural networks, allocating entire devices to these training tasks is wasteful and significantly delays any large model training.

Furthermore, the reliance on specialized hardware such as GPUs makes fine-grained resource sharing (i.e., training multiple networks on the same device) significantly harder than the typical examples in cloud systems. Unlike CPUs, the full virtualization of GPU resources is nascent~\citep{nvidia}. While modern GPU libraries support running multiple execution kernels in parallel, sharing resources using isolated kernels is not a mature solution in this setting. Many deep learning workloads are highly redundant, for example, the typical parameter tuning process trains the same model on the same data with small tweaks in hyperparameters or network architectures. In this setting, those parallel kernels would transfer and store multiple copies of the same training data on the device. This is analogous to the redundancy problems faced with conventional hypervisors running many copies of the same operating system on a single server~\cite{waldspurger2002memory}.

To avoid these pitfalls and provide efficient sharing, we need an approach that is aware of common I/O and computing processes among models that share a device.
We consider a scheme, packing models, where multiple static neural network architectures (e.g., ones that are typically used in computer vision) can be rewritten as a single concatenated network that preserves the input, output, and backpropagation semantics through a \texttt{pack} primitive. 
Not only does such concatenations facilitate partitioning of a single device it also allows us to synchronize data processing on GPUs and collapse common variables in the computation graph. It is often the case during hyperparameter tuning that the same model with various hyperparameter configurations are trained, and \texttt{pack} can feed a single I/O stream of training features to all variants of the model. 
In contrast, an isolated sharing way (e.g., training models isolatedly in sequence) may lead to duplicated work and wasted resources.

One of the surprising conclusions of this paper is that packing models together is not strictly beneficial. Counter-intuitively, certain packing policies can perform significantly worse than whole-device baselines--in other words, training a packed model can be slower than the sum of its parts (i.e., training these "parts" one by one). 
This paper studies the range of possible improvements (and/or overheads) for using \texttt{pack}. Further, we deploy \texttt{pack} to hyperparameter tuning and demonstrates that it can greatly improve the performance of hyperparameter tuning in terms of the time needed to find the best or the most promising model.


Our experimental results suggest: (1) There is a range of performance impact, spanning from 40\% faster execution to 10\% slower execution on a single GPU for packing two models over unpacking them for a single training step, and the improvement is scalable when packing more models. (2) The benefits of the \texttt{pack} primitive largely depend on a number of factors including memory capacity, chip architecture, neural network structure, batch size, and data preprocessing overlap. 
(3) There exists a trade-off between packing and unpacking when training multiple neural network models on limited resources. This trade-off is further complicated by architectural properties that might make a single training step bounded by computation (e.g., backpropagation is expensive) or data transferring (e.g., transferring training batches to GPU memory). (4) The \texttt{pack} primitive can speedup hyperparameter tuning by up to 2.7$\times$.

\section{Background}
\label{sec:bg}
\sloppy
First, we motivate and illustrate the \texttt{pack} primitive.

\subsection{Motivation}
\begin{figure}[htbp]
    \centering
    \vspace{-5pt}
    \includegraphics[width=0.9\columnwidth]{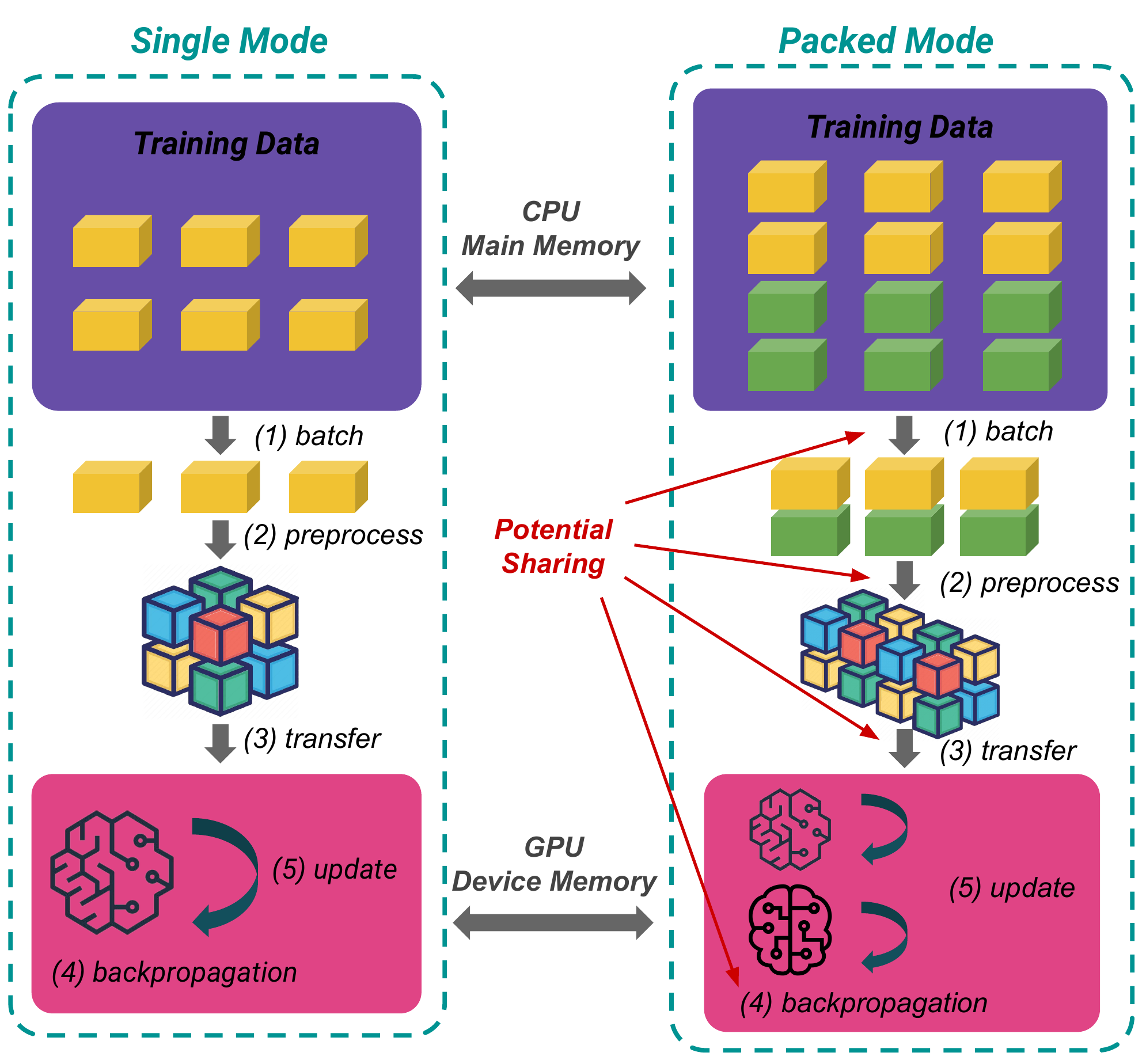}
    \vspace{-8pt}
    \caption{The dataflow of a training step in the single mode v.s. the packed mode. The training data resides in main-memory and is copied over to the device in batches during each training step resulting in a backprogation computation and then a parameter update. By synchronizing the dataflow, the packed mode can reuse work when possible.}
    \vspace{-10pt}
    \label{fig:teaser}
\end{figure}

Figure \ref{fig:teaser} demonstrates the typical data flow in neural network model training with stochastic gradient descent (or related optimization algorithms).
We use the term \textit{host} to describe CPU/Main-Memory/Disk hierarchy and \textit{device} to refer to the GPU/Device Memory. 
In this setup, all of the training data resides on the host.
Considering the typical training setup on the left side, a batch of data is taken from the host and copied to the device.
Additionally, it is common in machine learning (especially in Computer Vision) that this data is preprocessed before it is transferred.
Then, on the device, the execution framework calculates a gradient using backpropagation.
Finally, using the results from the backpropagation, the model is updated.

In the typical "multiple-tasks-single-device" mode, resource sharing is often temporal--where one training task uses the whole GPU first and then switches full control to another task. Resource sharing in single mode is wasteful if the models are small and there is sufficient GPU memory to fit both models on the device simultaneously.
The right side of Figure \ref{fig:teaser} motivates a different solution. It allows multiple models to be placed on a single GPU. This packed mode could bring some potential benefits. Suppose, we are training two models on the same dataset to test if a small tweak in neural network architecture will improve performance. The same data would have to be copied and transferred twice for training. If the system could pack together models when compatible in size, then these redundant data streams can be fused together to improve performance. 


\subsection{Basic Framework API}
We desire a framework that can pack models together and jointly optimize their respective computation graphs when possible to reduce redundant work. We assume that we have access to a full neural network description, as well as the weights of the network.
Each training task is characterized by four key traits: (a) \textit{Model.} A computation graph architecture of the model with pointers to the input and output, equipping with some training hyperparameters (e.g. learning rate, optimizer, etc.) and assigning to a logical name that is uniquely identified. (b) \textit{Device.} The target device to be used for placing and training models. (c) \textit{Batch Size.} The batch size used in the training process, where each batch refers to the size of input data used in a single training step. (d) \textit{Training Step.} The number of steps to train the model, which is also relevant to the number of epochs since typically one epoch consists of numerous steps.  


Our objective is the following isolation guarantee: given these four traits, our framework will train the models in a fine-grained way but preserve the accuracy as if the training tasks were trained isolatedly and sequentially on a dedicated device. No action that the framework takes should affect training accuracy. Such a framework requires three basic primitives \texttt{load}, \texttt{free}, and \texttt{pack}. Users should be able to interact with our framework without worrying about exactly how the resources are allocated and on which devices the models are placed.

The primitives \texttt{load} and \texttt{free} can "copy in" and "copy out" models. Given a device name and model, \texttt{load} places the model on the device:
\begin{verbatim}
load(model, device)
\end{verbatim} 

Given a device name and model, \texttt{free} retrieves the model and frees the resources taken by the model:

\begin{verbatim}
checkpt = free(model, device)
\end{verbatim} 

State-of-the-art neural network training algorithms have additional state as a part of the optimizer.
This state is stored with the model (see our experiments on computer vision models with optimizers in Section \ref{sec:profiling}). Then, the API provides the primitive \texttt{pack} for packing. Suppose, we have two neural network models:

\begin{verbatim}
output1 = nn1(input1)
output2 = nn2(input2)
\end{verbatim} 

The \texttt{pack} primitive combines both models into a new neural network by concatenating the output layers:

\begin{verbatim}
[output1 output2] = packed_nn([input1 input2])
\end{verbatim} 

This packing operation is fully differentiable and preserves the correctness semantics of the two original networks. Crucially this allows the execution layer to process inputs simultaneously. 

Thus, the models can be jointly trained using \texttt{pack}. The training steps have to be synchronized in the sense that the models are differentiated and updated at the same time. This synchronization leads to a complex performance trade-off, if the models are too different the device may waste effort stalling on one model while either updating or differentiating on the other. This means that training a packed model may be significantly slower than sequentially training each constituent model in it. However, the overheads from stalling may be counteracted by the benefits of reducing redundant computation. Navigating this complex trade-off space is the motivation for this study, and we seek to understand under what conditions is \texttt{pack} beneficial.



\section{Implementation}
\label{sect:tf-impl}
 We build a prototype system on top of TensorFlow to implement the above framework APIs and take image classification as our motivating application in our implementation, but the idea of \texttt{pack} is generally compatible with other platforms and applications.



\subsection{Packing}
\texttt{Pack} is a lossless operation that concatenates the outputs of two or more neural network models. Since it is lossless, it preserves the forward and backward pass semantics of the model. The basic operation can be written as packing multiple output variables, as illustrated in the following example:

\begin{verbatim}
mlp_out = #reference to mlp output
resnet_out = #reference to resnet output
densenet_out = #reference to densenet output
packed_out = pack([mlp_out resnet_out densenet_out])
\end{verbatim}

This \texttt{packed\_out} can be thought of as a new neural network model that takes in all input streams (even possible different input data types) and outputs a joint prediction.
Thus, we can do everything to a packed model that we could do to a single neural network.
The packed model can be differentiated and the model parameters can be updated iteratively.
The model can be placed on a device, such as GPU or TPU, as a single unit.

While this gives us scheduling flexibility, there is a major caveat.
By packing the models together, we create an artificial synchronization barrier. If one of the models is significantly more complex than the others, it will block progress.
Likewise, if one of the operations saturates the available compute cores, progress will stall as well. 
Naive packing leads to a further issue where the input batch has to be synchronized in dimension as well (each model is differentiated or evaluated the same number of times). Therefore, without further thought, the scope of \texttt{pack} is very narrow.

\subsection{Misaligned Batch Sizes}
\label{subsubsec:mbs}

Requiring that all packed models have the same batch size is highly restrictive, but we can relax this requirement.
Our method is to rewrite the packed model to include a dummy operation that pads models with the smaller batch size to match the larger ones in dimension.
The \texttt{pad} primitive is exploited for packing models with different batch sizes. The original models are packed and trained based on the batch with the largest size, but the batches for the models with smaller batch sizes will be padded. 
During training and inference, the padding is sliced.

\vspace{-2pt}
\begin{figure}[htbp]
    \centering
    \includegraphics[width=0.8\columnwidth]{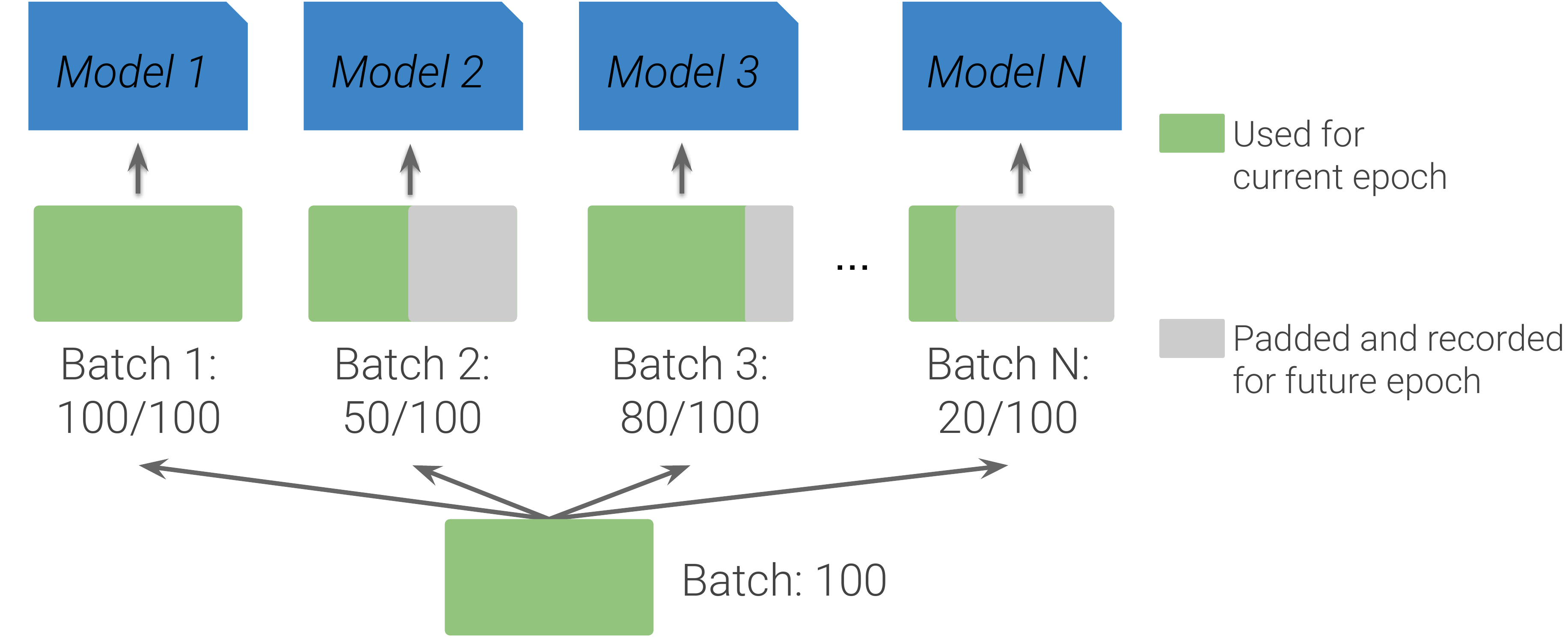}
    \vspace{-5pt}
    \caption{All the models share the batch input stream, each batch is padded and sliced for training the packed model.}
    \vspace{-5pt}
    \label{fig:packed-diff-batch}
\end{figure}
\vspace{-3pt}

As depicted in Figure \ref{fig:packed-diff-batch}, there are a set of original models with various batch sizes, the largest training batch size (i.e., $100$) is selected and fed to the packed model for a single training step accordingly. Then, the batch will be replicated for $n$ original models in the packed model. The model $1$ takes the entire batch, whereas the replicated batches $2, \cdots, N$ are sliced to match the models' requirement. Thus, all the models can be trained together.

Simply slicing may result in statistical inefficiency since only a fraction of the entire dataset is used during each epoch for the models with smaller batch size. To address this issue, we track the progress of each model individually to ensure that there is no loss in training dataset. Assuming we train the packed model in Figure \ref{fig:packed-diff-batch} for one epoch. When model 1 finishes training and is unloaded, model 2 achieves $50\%$ progress and uses $50\%$ of the training dataset, model 3 has been training using $80\%$ dataset, and so do the other models (dataset usage is recorded for all models). Then, the packed model takes the current largest batch size (i.e. $80$ from model 3) and uses the rest training dataset of model 3 to train the packed model. Due to slicing, it is obvious that unused training dataset of model 3 are included in the unused dataset of other models. The process continues until all models are trained completely and thus no training data is missing. 

\subsection{Misaligned Step Counts}
\label{subsubsec:msc}

Another issue with synchronization is that different models may need to be trained for a different number of steps. Even if all of the models are the same, this can happen if the user is trying out different batch sizes. 

\begin{figure}[htbp]
    \centering
    \includegraphics[width=0.9\columnwidth]{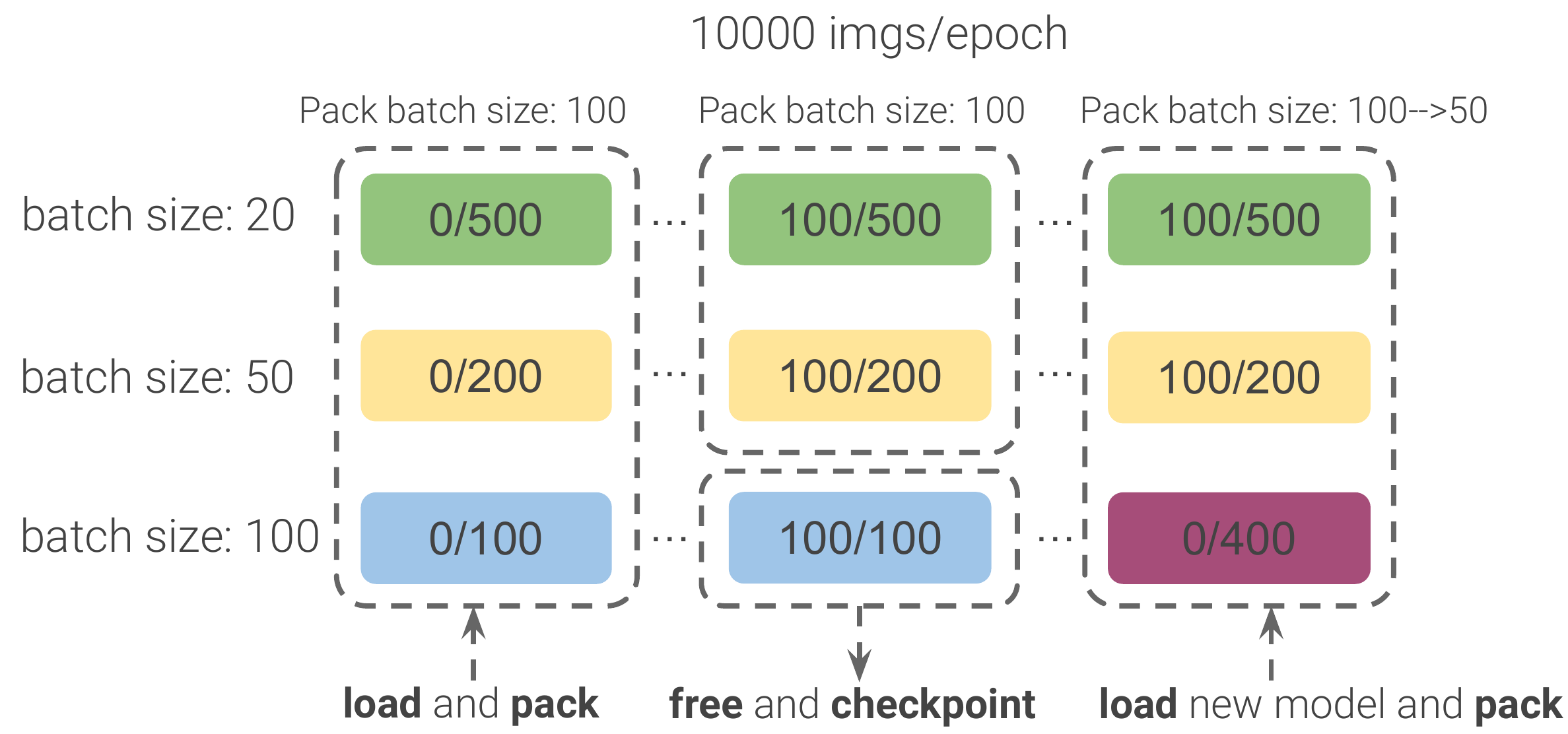}
    \vspace{-5pt}
    \caption{Early finished model is freed and checkpointed, new model is packed with the others for further training.}
    \vspace{-6pt}
    \label{fig:packed-diff-step}
\end{figure}

We use \texttt{load} and \texttt{free} to address this issue. As demonstrated in Figure \ref{fig:packed-diff-step}, we train three models with batch size $20$, $50$, and $100$ for one epoch using $10,000$ images and labels, and they requires 500 steps, 200 steps, and 100 steps. We packed them for training, and when the model with $100$ steps is finished, it is freed and checkpointed. Then, a new model can be loaded and packed with the incomplete models to continue training. This brings an overhead of loading but freeing the device memory for further training.

\subsection{Eliminating Redundancy}
\texttt{Pack} forces synchronization, which means that dimensional differences between the models or training differences between the models can lead to wasted work. However, \texttt{Pack} can allow the system to eliminate redundant computations and data transfers. Consider a hyperparameter tuning use case where we are training the same network with a small configuration tweak on the same dataset:

\begin{verbatim}
nn_conf1_out = nn_conf1(input1)
nn_conf2_out = nn_conf2(input2)
\end{verbatim}

In this case, \texttt{input1} and \texttt{input2} refer to the same dataset. We can avoid transferring the batch multiple times by symbolically rewriting the network description to refer to the same input:

\begin{verbatim}
nn_packed_out = pack([nn_conf1_out nn_conf2_out])
[output1 output2] = nn_packed_out(input)
\end{verbatim}

The potential upside is significant as it reduces the amount of data transferred along a slower I/O bus. Furthermore, eliminating redundant computation goes beyond identifying common inputs. Preprocessing is a common practice for machine learning training tasks. The preprocessing operations (e.g., data augmentation, image decoding) happen before training and can actually dominate the total execution time of some models. When packing models that take the same preprocessing, the \texttt{pack} primitive can fuse the steam processing and eliminate redundant tasks. This idea can be extended if multiple models have fixed featurization techniques or leverage the same pretrained building blocks.



\section{Profiling Model Packing}
\label{sec:profiling}
As it stands, model packing leads to the following trade-offs. Potential performance improvements include: (1) eliminating redundant data transfers when models are trained on the same dataset, (2) combining redundant computations including preprocessing, (3) performing computations (forward and back propagation) in parallel if and when possible. On the other hand, the potential overheads include (a) models that dominate the device resources and block the progress of the others, (b) overhead due to misaligned batch sizes, and (c) overhead due to loading and unloading models with a differing number of training steps.

This section describes a series of experiments that illustrate when (what architectures and settings) packing is most beneficial. 

\subsection{Profiling Setup}
\label{subsec:setup}
Our server is 48-core Intel Xeon Silver 4116@2.10GHz with 192GB RAM, running Ubuntu 18.04. The GPU is NVIDIA Quadro P5000. Our evaluation uses 4 models: Multilayer Perceptron with 3 hidden layers (MLP-3), MobileNet \cite{DBLP:conf/cvpr/SandlerHZZC18}, ResNet-50 \cite{DBLP:conf/eccv/HeZRS16}, and DenseNet-121 \cite{DBLP:conf/cvpr/HuangLMW17} -- with all models implemented in TensorFlow 1.15. The default training dataset is $10,000$ images from ImageNet \citep{ILSVRC15} and the required input image size of each batch is $224 \times 224$ which is commonly used. Batch sizes start from $32$ and goes up to $100$ in the experiments \citep{DBLP:series/lncs/Bengio12}. 

In our experimental methodology, the first training step is always omitted for measurement due to the CUDA warm-up issue, and the measurement of the single step excluded loading time. We only measure the loading cost for investigating whether it dominates the performance (middle column in Figure \ref{fig:exp-micro-gpu}). So, this measurement is orthogonal to any pipelining that might happen at a different level of abstraction. The results in the paper are averaged over $5$ independent runs.

\subsection{Profiling Metrics}
We evaluate the \texttt{pack} primitive against three performance metrics defined as follows.

\noindent \textbf{\underline{Improvement: }} We measure the time of a single training step of the packed model. Since one training epoch can be treated as a series of repeating training steps and a complete training process is made with multiple epochs, the single training step measurement can be used to estimate the overall training time.
We denote the step time as $T_s$ and assume that there are $n$ models (model $1, \cdots, n$), and we compare the time of a single training step in packed and sequential mode. We first train models $1, \cdots, n$ isolatedly and sequentially, and measure the time of a single training step:
\vspace{-0.1cm}
\begin{equation}
T_s(Seq) = T_s(\text{Model } 1) + \cdots + T_s(\text{Model } n)
\vspace{-0.1cm}
\end{equation}
Then, we pack these models for training and measure the single training step, which is defined as follows:
\vspace{-0.1cm}
\begin{equation}
T_s(Pack) = T_s(Pack(\text{Model } 1, \cdots, n))
\vspace{-0.1cm}
\end{equation}
Thus, we define the improvement metric as follows:   
\vspace{-0.1cm}
\begin{equation}
\label{eq:impv}
    IMPV = \frac{T_s(Seq) - T_s(Pack)}{T_s(Seq)}
\vspace{-0.1cm}
\end{equation}
The improvement metric can quantify the benefits brought by \texttt{pack} primitive, and comparing $IMPV$ of various training setups can identify performance bottlenecks.

\noindent \textbf{\underline{Memory: }} Fine-grained resource sharing (e.g., training multiple models together on a single device) requires sufficient device memory, thus measuring the memory usage of the packed model can provide insights for scheduling different models given a specific device memory capacity. 
We evaluate the peak of memory usage over the training epoch. This is because if the usage peak is over the GPU memory capacity, the training process will be terminated due to a GPU memory error. We measure the allocated memory and not the active memory used. 

\noindent \textbf{\underline{Switching Overhead: }} Training the models isolatedly and sequentially on a single device can bring an additional switching overhead. For example, the GPU has to unload the old model and the associated context and then load the new models and prepare the context. \texttt{Pack} significantly reduces such overhead since packing models suffer from model switching less often (multiple models can be trained together given enough GPU memory so that loading and unloading operations can be avoided). The switching overhead is measured through the following method: We train $n$ models isolatedly and sequentially for one epoch and capture the training time, which is denoted as $T_e(Seq)$. Then we train $n$ models individually and denote $T_e(Model)$ as the training time of one epoch for each model. Thus, the switching overhead of training $n$ models is defined as:
\begin{equation}
    SwOH(n) = T_e(Seq) - T_e(\text{Model 1}) \cdots - T_e(\text{Model n})
\end{equation}
However, we hypothesize that the overhead amortizes over an entire training procedure. This is because $SwOH$ depends on the number of models instead of the number of training steps and epochs. Since training a model usually involves numerous training steps and many training epochs, compared with much longer training time, the switching overhead is minor (section \ref{sect:swoh}).

\subsection{Improvement}
\begin{figure}[h]
    \centering
    \vspace{-10pt}
    \subfigure[Increasing number of models]{\label{fig:exp-gpu-model-improve}\includegraphics[width=0.49\columnwidth]{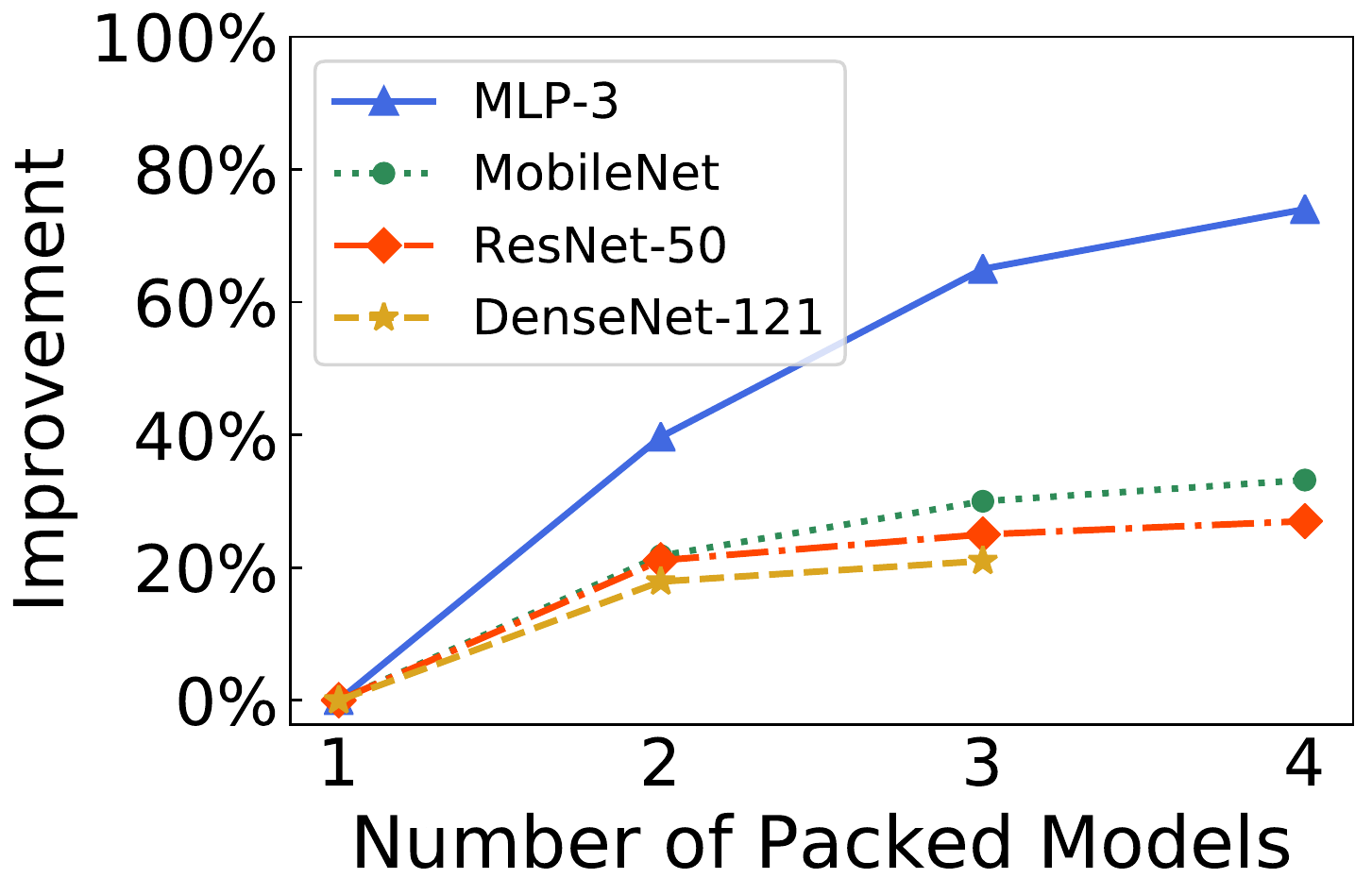}}
    \subfigure[Increase batch size]{\label{fig:exp-gpu-bs-improve}\includegraphics[width=0.49\columnwidth]{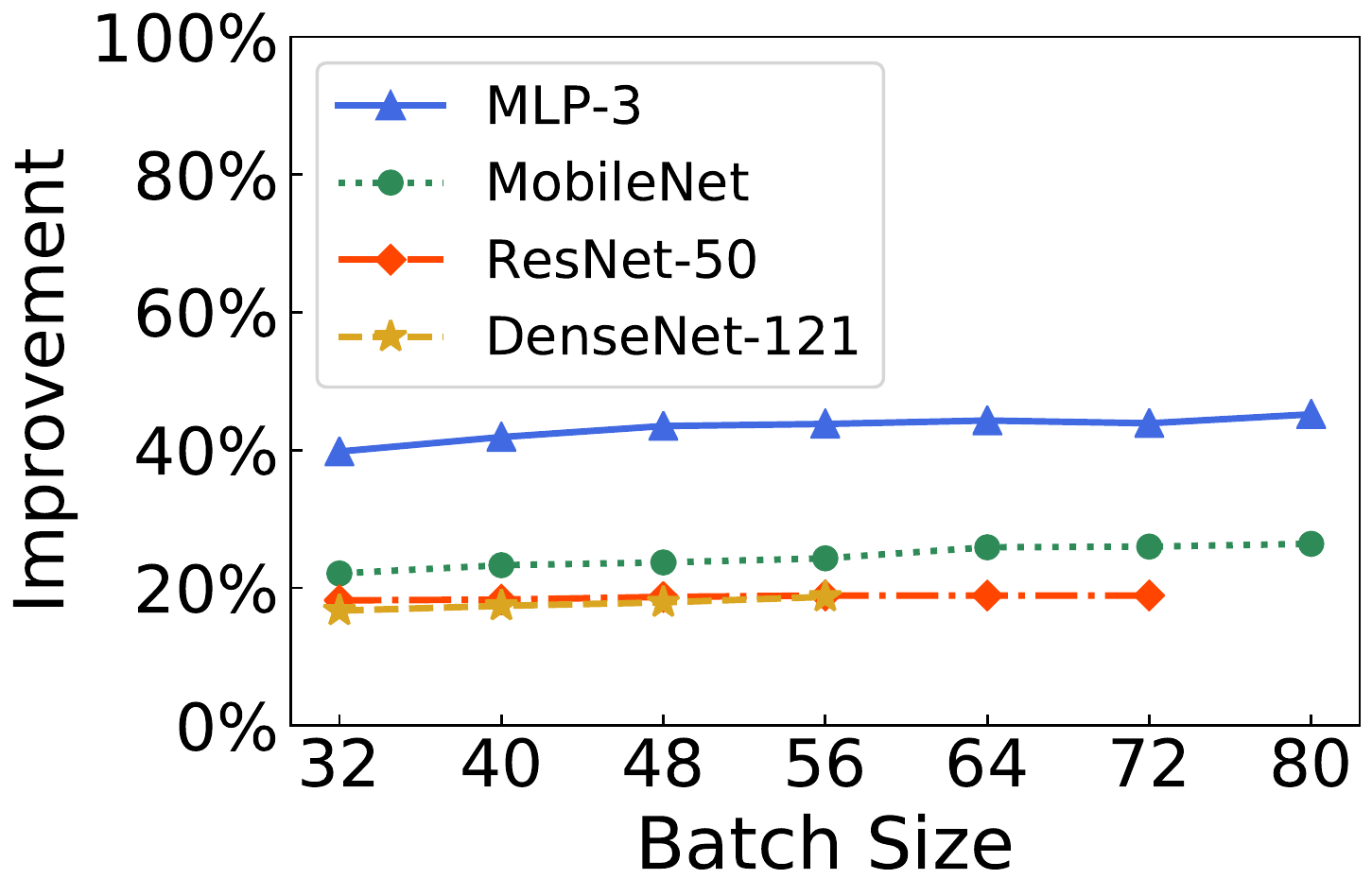}}
    \vspace{-12pt}
    \caption{\textit{Improvement} of packing models when increasing number of models and batch size on GPU. Y-axis indicates the reduction in training time compared to sequential execution (Eq. \ref{eq:impv}).}
    \vspace{-13pt}
    \label{fig:exp-gpu-improve}
\end{figure}

We evaluate packing performance as a function of batch size and the number of models. Figure \ref{fig:exp-gpu-model-improve} shows that as the number of packed models increases so do the relative benefits until the resources are saturated.
The line of DenseNet-121 ends early because that packing four DenseNet-121 takes too much GPU memory and results in an Out-Of-Memory (OOM) issue. 
However, the potential for resource savings is significant. 
If one is training multiple MLP models, there can be up to an 80\% reduction in training time.
In short, \emph{it is wasteful to allocate entire devices to small models.}

Figure \ref{fig:exp-gpu-bs-improve} illustrates the relationship between batch size and relative improvement when packing two models. The lines of ResNet-50 and DenseNet-121 both end early because the OOM issue emerges when the batch size goes to $80$ and $64$ respectively. These models are mostly GPU-compute bound. 
Increasing the batch size has a negligible improvement in time even if the packing setup can combine the data transfer.
We will see that this story gets more complicated when considering preprocessing.

\subsection{Memory Usage}
We track the GPU memory usage of training individual models and packed models with different batch sizes for one epoch. We particularly care about the memory peak and whether it is beyond the memory capacity. 

\begin{figure}[htp]
    \centering
    \vspace{-5pt}
    \includegraphics[width=\columnwidth]{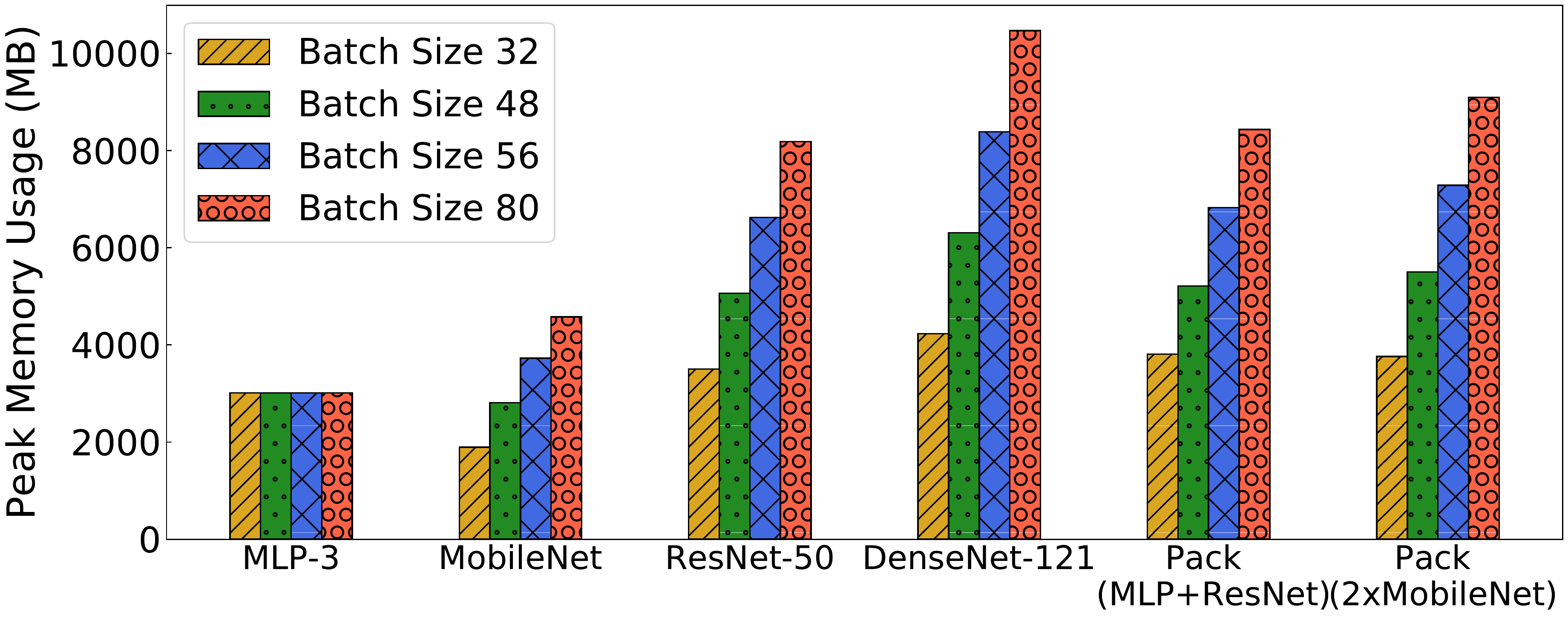}
    \vspace{-22pt}
    \caption{GPU memory peak of different models}
    \vspace{-10pt}
    \label{fig:exp-gpu-mem}
\end{figure}

As depicted in Figure~\ref{fig:exp-gpu-mem}, for convolutional neural networks like ResNet, MobileNet, and DenseNet, the GPU memory usage is proportional to the batch size as more intermediate results will be stored as batch size increases. Similarly, when packing two models the GPU memory usage is the sum of memory usage. However, the GPU memory usage peak of MLP-3 model maintains the same as the batch size goes up. This is mainly due to two reasons: (1) we find that for simple models TensorFlow's greedy memory allocation policy over-allocates more GPU's memory when the actual usage is lower than a specific threshold; (2) the majority of computations for MLP-3 are dot products and are placed on CPU by TensorFlow and do not occupy much GPU memory. More specifically, without any annotations, TF automatically decides whether to use the GPU or CPU for an operation \cite{tfbasics} (we also used the TF profiler to trace the training process and found the majority of operations in MLP-3 are placed on the CPU). Thus, GPU memory usage of single MLP-3 remains the same due to the pre-allocation.

\subsection{Switching Overheads}
\label{sect:swoh}
We profile the switching overhead of two models to illustrate how much the overhead can accumulate as more models are trained (the time \texttt{pack} can save). 

As shown in the Table \ref{tab:swoh}, albeit the accumulation, the switching overhead (using Eq. \ref{fig:exp-gpu-improve}) is minor compared to the overall training time and it is even negligible when more epochs are involved in a training process. This also confirms our hypothesis of the switching overhead.


\begin{table}[htp]
\vspace{-4pt}
\newcolumntype{?}{!{\vrule width 2pt}}
\onehalfspacing
\small
\centering
\begin{tabular}{c|c|c|c|c}
\Xhline{3\arrayrulewidth}
\hline
\hline
& Model & $T_e(Seq)$ & $T_e(Model)$ & $SwOH(2)$ \\
\hline
\multirow{4}{*}{GPU} & MLP-3 & 133s & 61s & 11s \\ 
& MobileNet & 227s & 107s & 13s  \\
& ResNet-50 & 274s & 130s & 14s \\ 
& DenseNet-121 & 305s & 144s & 17s \\ 
\hline
\hline
\Xhline{3\arrayrulewidth}
\end{tabular}
\caption{$SwOH$ of training two models sequentially}
\vspace{-20pt}
\label{tab:swoh}
\end{table}

\begin{table*}[htbp]
\newcolumntype{?}{!{\vrule width 2pt}}
\onehalfspacing
\centering
\small
\begin{tabular}{P{1.6cm}|P{1cm}|p{14cm}}
\Xhline{3\arrayrulewidth}
\hline
\hline
\textbf{Factor} & \textbf{Config} & \textbf{Description} \\ 
\hline
\multirow{2}{*}{\parbox[c][1.3cm]{1.6cm}{\centering Model}} & Same & Packing two same models\\
\cline{2-3}
& \multirow{1}{*}{\parbox[c][0.8cm]{1cm}{\centering Different}} & Packing two different models. To figure out more configurations, we evaluated MLP-3 vs. MobileNet, MobileNet vs. DenseNet-121, ResNet-50 vs. MobileNet, and DenseNet-121 vs. ResNet-50. \\
\hline
\multirow{2}{*}{Training Data} & Same & All packing models take the same training batch data \\
\cline{2-3}
& Different & All packing models take the different training batch data. \\
\hline
\multirow{2}{*}{Preprocess} & \multirow{1}{*}{\centering Yes} & Preprocessing is included in each training step. Training batch are raw image (e.g., JPEG), transferring from disk to GPU. \\
\cline{2-3}
& \multirow{1}{*}{\centering No} & Preprocessing is excluded in each training step. Training batch are preprocessed and formatted before transferring to GPU.\\
\hline
\multirow{2}{*}{Optimizer} & Same & Two models use the same optimizer for single training step, e.g., both of them use Momentum optimizer. \\
\cline{2-3}
& \multirow{1}{*}{\centering Different} & Two models use the different optimizer for single training step, e.g., one uses Momentum, the other uses SGD. \\
\hline
\multirow{2}{*}{Batch Size} & Same & Two models take the same batch size for single training step, e.g., both of them take 32 batch size. \\
\cline{2-3}
& \multirow{1}{*}{\parbox[c][0.25cm]{1.1cm}{\centering Different}} & Two models take the different batch sizes for single training step, e.g., one is 32 batch size, the other is 50 batch size.\\
\hline
\hline
\Xhline{3\arrayrulewidth}
\end{tabular}
\caption{Model configurations for ablation study} \label{tab:micro-conf}
\vspace{-12pt}
\end{table*}

\subsection{\texttt{Pack} vs CUDA Parallelism}
Current NVIDIA GPUs support executing multiple CUDA kernels in parallel at application level. Thus, we conduct an experiment under the same environment as we used in the paper to train models in parallel at the CUDA GPU kernel. We run multiple simultaneous training processes on TensorFlow. We evaluate this method in the experiments where two processes are boosted at the same time to train the same models (MLP, MobileNet, ResNet, DenseNet) with the same optimizer and same batch size (ranging from $32$ to $100$).

Although CUDA supports it, our results show that it is not an efficient technique. When the models train on the same data, parallel training in isolated kernels leads to duplicated I/O and duplicated data in memory. In the image processing tasks that we consider, the training data batch takes up a substantial amount of memory. We find that in all but the simplest cases lead to an OOM error: "failed to allocate XXX from device: \texttt{CUDA\_ERROR\_OUT\_OF\_MEMORY}". We also find similar results when the models train on \emph{different} data---as there is duplicated TensorFlow context information in each of the execution kernels. This error happens all the above experiments except packing the MLP model (due to its lightweight size).

Even with the MLP model, the \texttt{pack} primitive shows benefits at scale. For instance, the training time of a single step based on CUDA parallelism is $184$ms for both two processes and the packing method takes $200$ms. However, as the batch size is increased to $100$, the former one takes $1660$ms, while the latter one costs $1500$ms. We interpret these numbers as an indication that the \texttt{pack} primitive incurs smaller context overhead over the native CUDA parallelism at application level.


\begin{figure*}[htbp]
    \centering
    \subfigure{\label{fig:exp-gpu-model-mlp}\includegraphics[width=0.41\columnwidth]{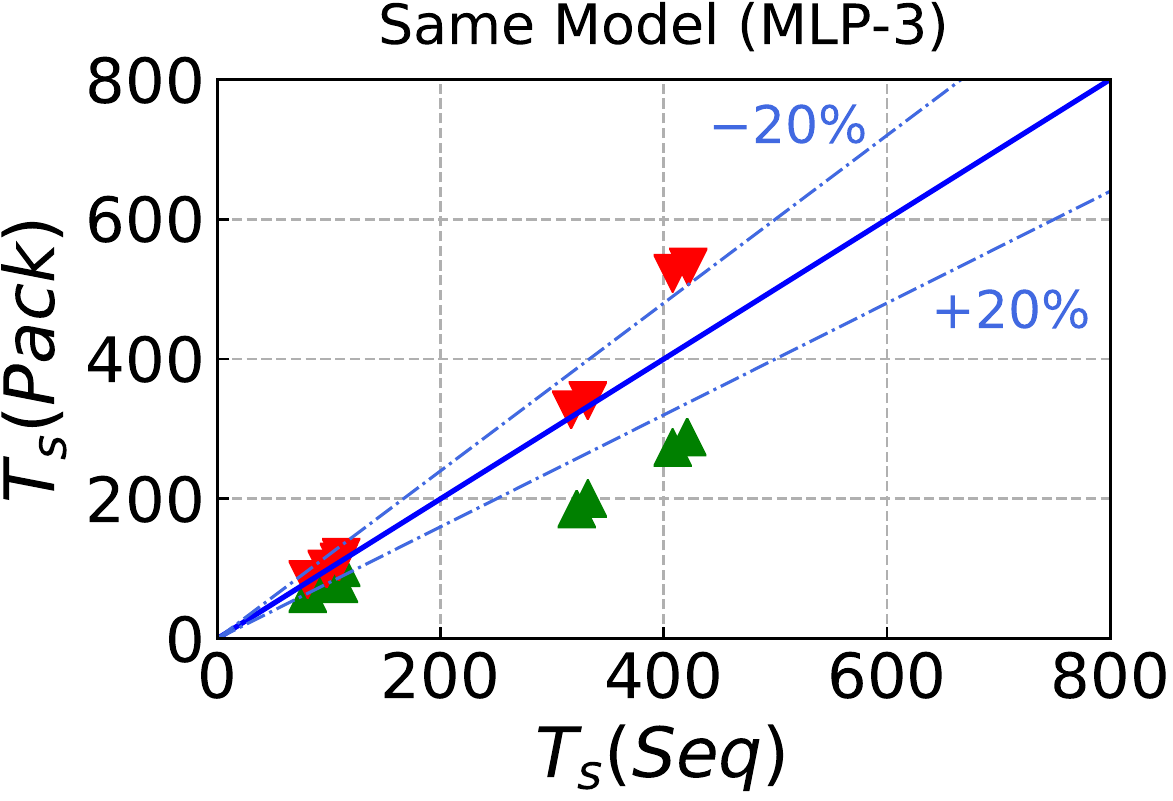}}\hspace{-3pt}\vspace{-1pt}
    \subfigure{\label{fig:exp-gpu-data-mlp}\includegraphics[width=0.41\columnwidth]{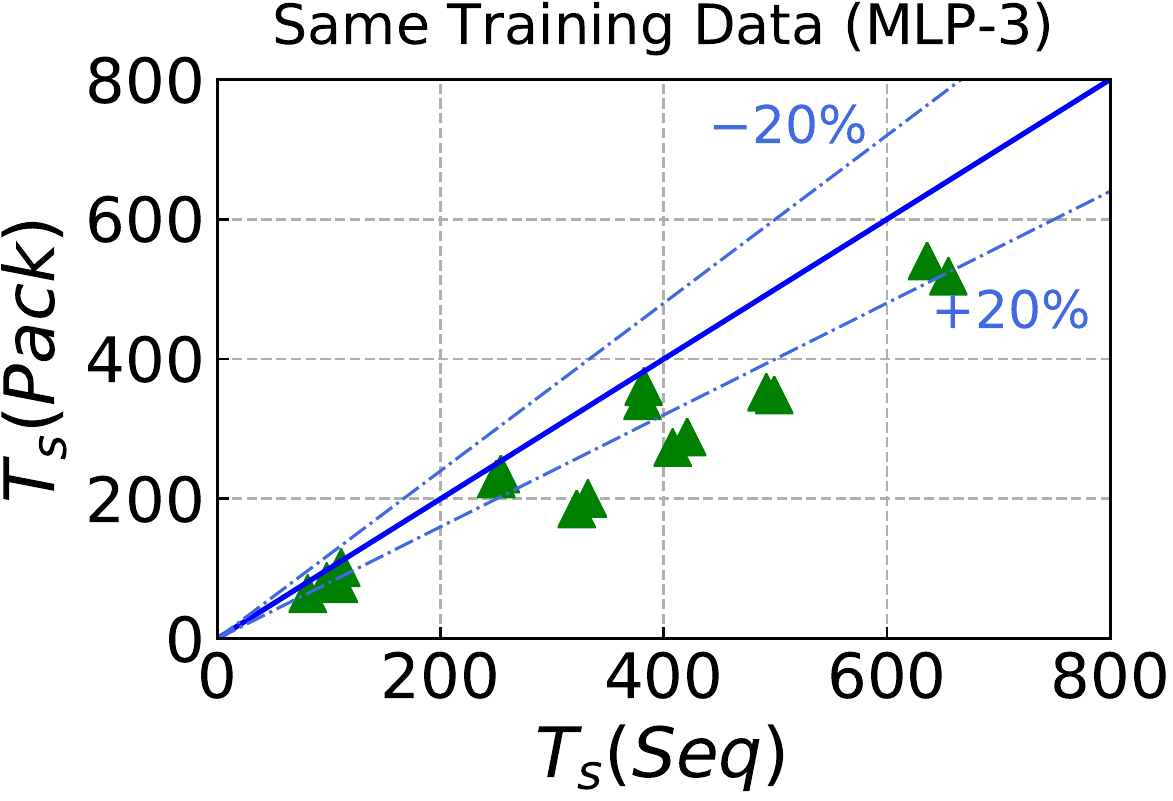}}\hspace{-3pt}\vspace{-1pt}
    \subfigure{\label{fig:exp-gpu-prep-mlp}\includegraphics[width=0.41\columnwidth]{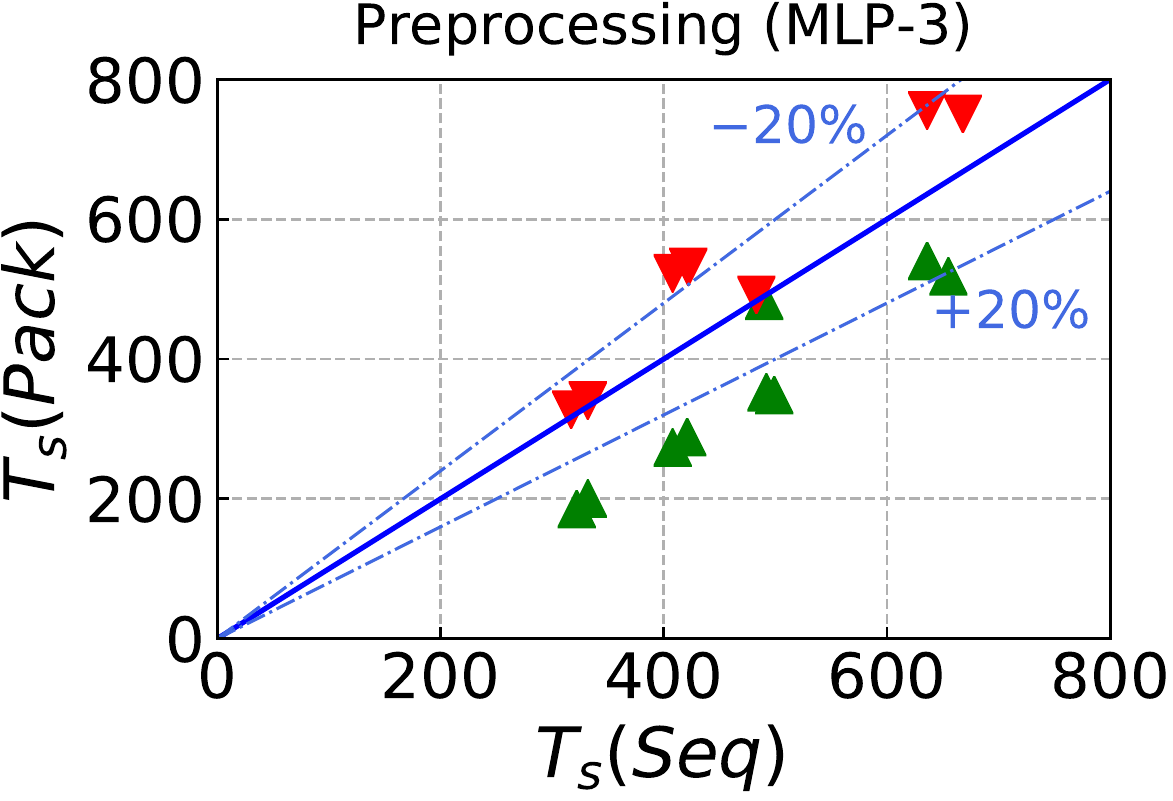}}\hspace{-3pt}\vspace{-1pt}
    \subfigure{\label{fig:exp-gpu-opt-mlp}\includegraphics[width=0.41\columnwidth]{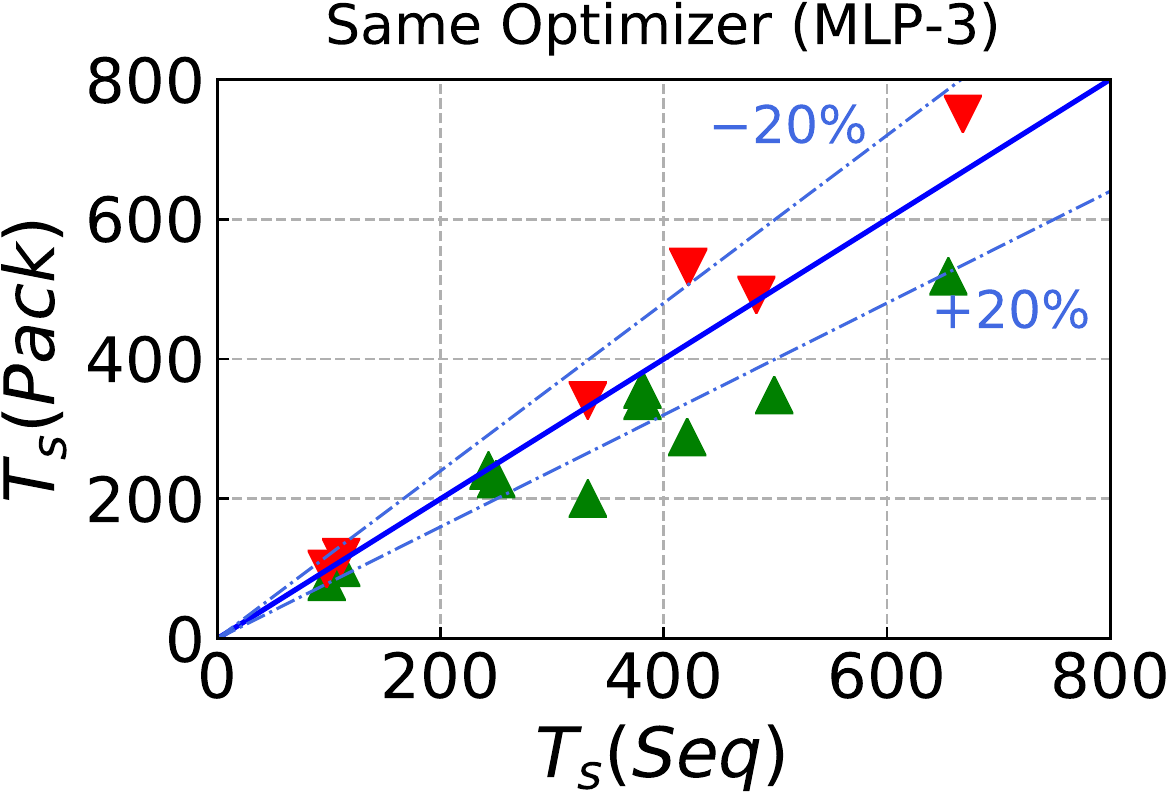}}\hspace{-3pt}\vspace{-1pt}
    \subfigure{\label{fig:exp-gpu-bs-mlp}\includegraphics[width=0.41\columnwidth]{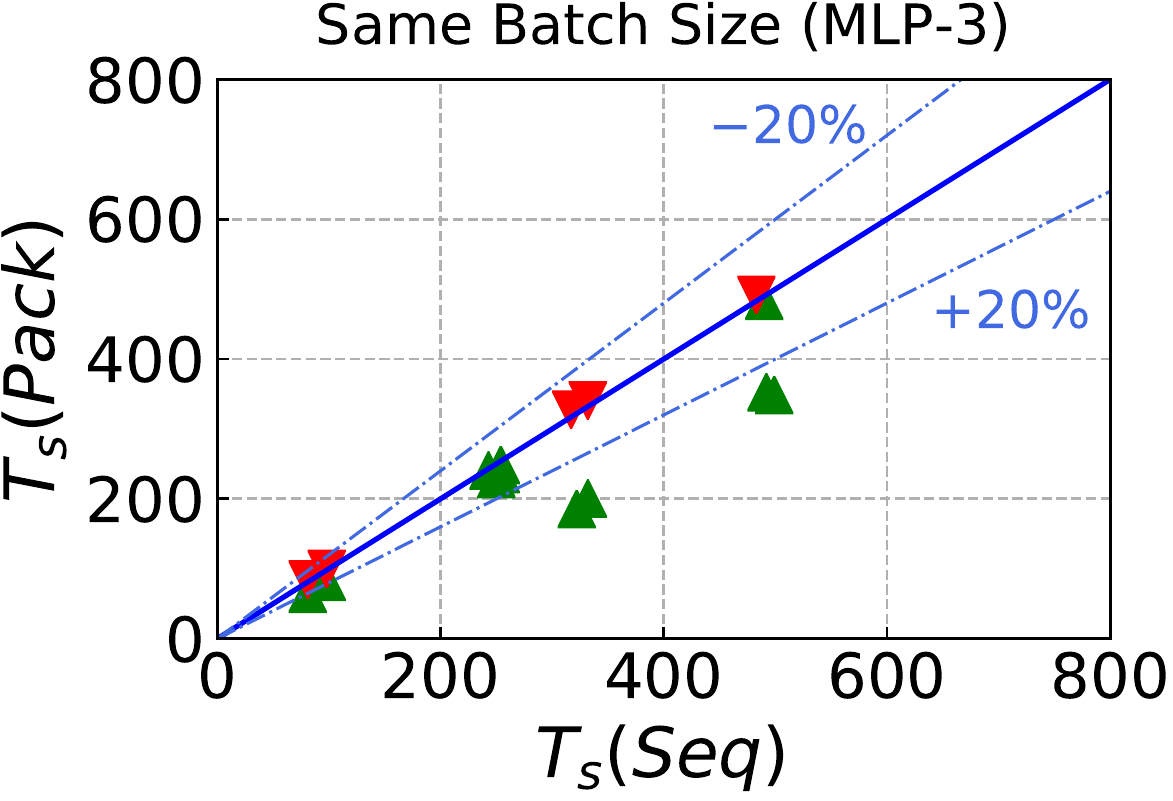}}\hspace{-3pt}\vspace{-1pt}
    \subfigure{\label{fig:exp-gpu-model-mobile}\includegraphics[width=0.41\columnwidth]{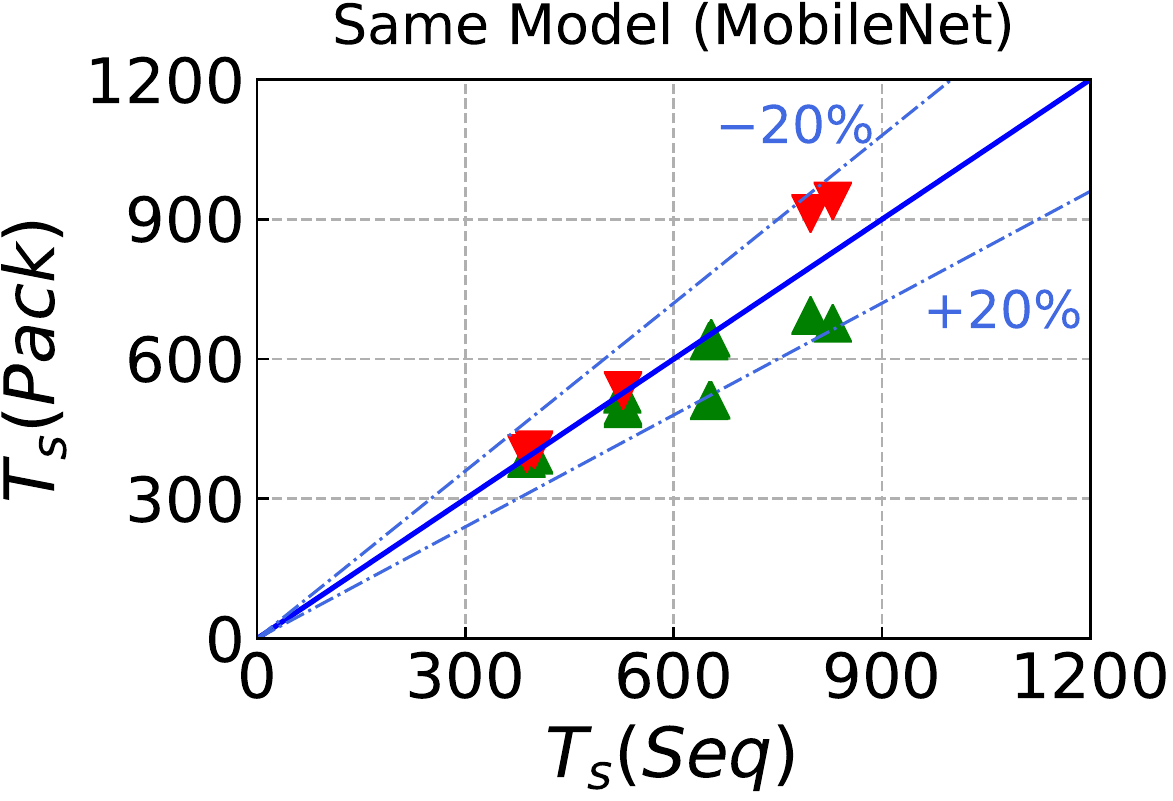}}\hspace{-3pt}\vspace{-1pt}
    \subfigure{\label{fig:exp-gpu-data-mobile}\includegraphics[width=0.41\columnwidth]{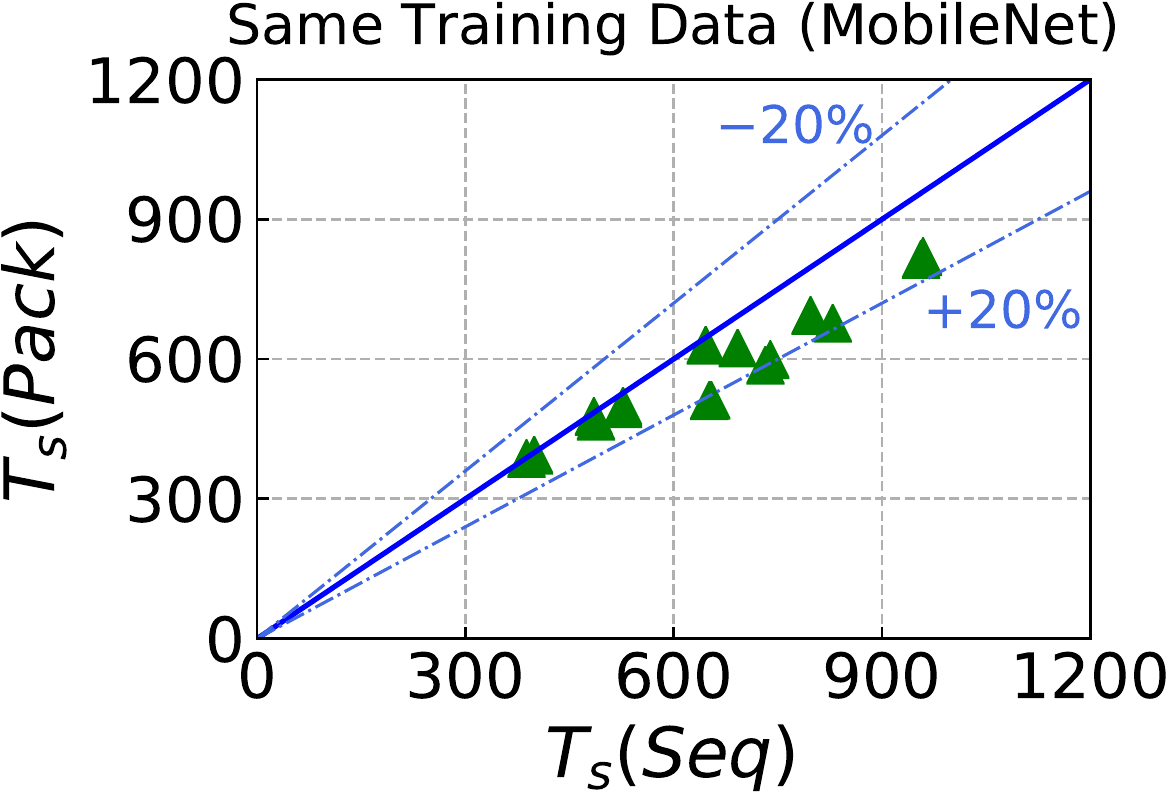}}\hspace{-3pt}\vspace{-1pt}
    \subfigure{\label{fig:exp-gpu-prep-mobile}\includegraphics[width=0.41\columnwidth]{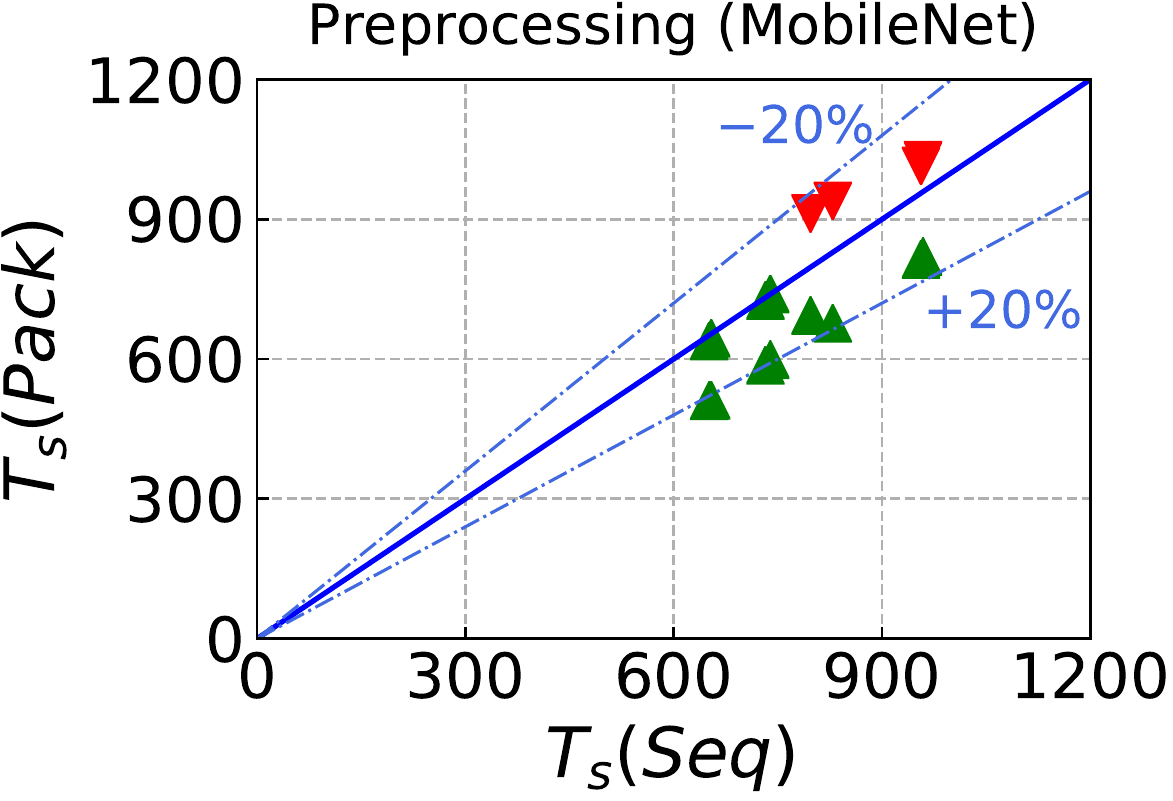}}\hspace{-3pt}\vspace{-1pt}
    \subfigure{\label{fig:exp-gpu-opt-mobile}\includegraphics[width=0.41\columnwidth]{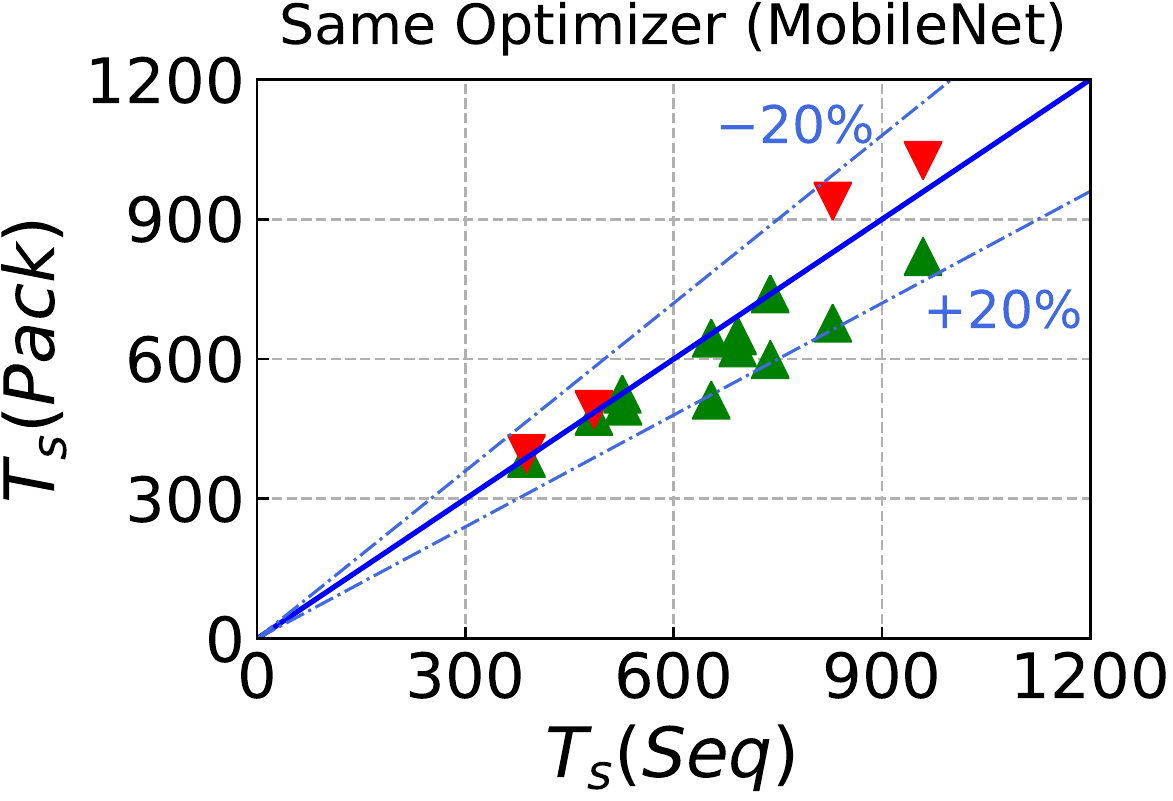}}\hspace{-3pt}\vspace{-1pt}
    \subfigure{\label{fig:exp-gpu-bs-mobile}\includegraphics[width=0.41\columnwidth]{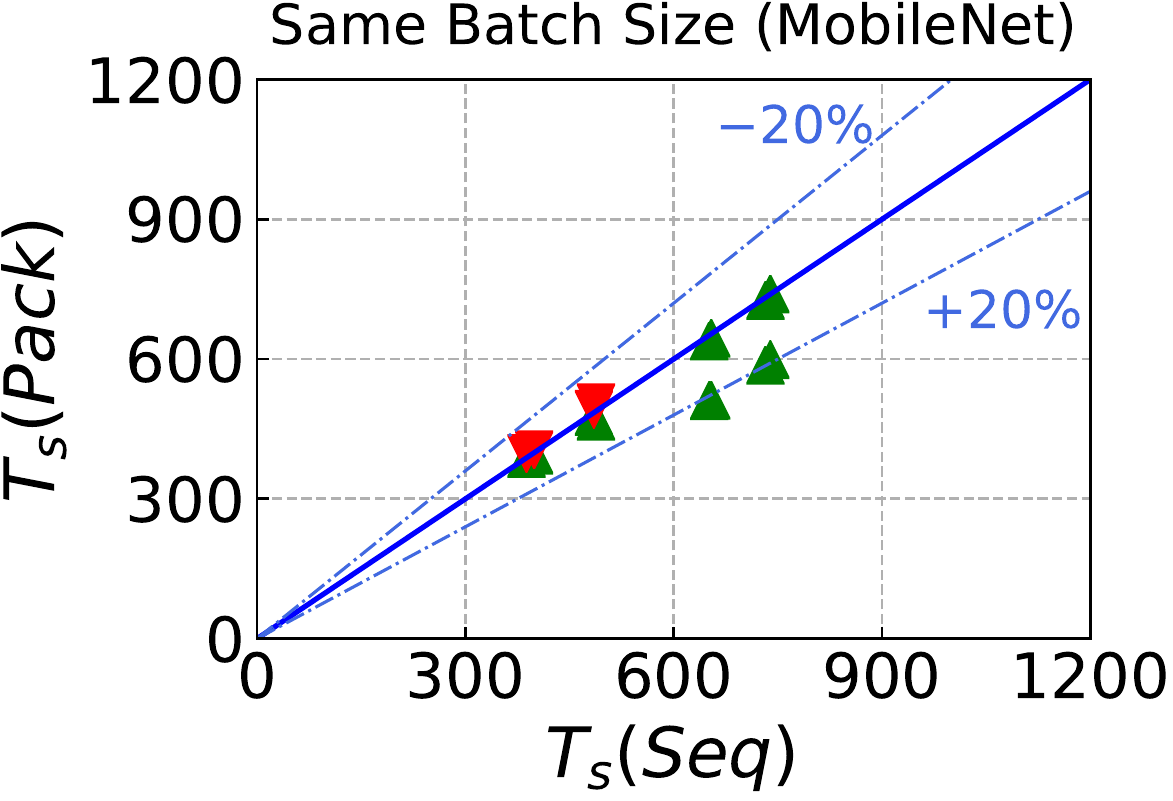}}\hspace{-3pt}\vspace{-1pt}
    \subfigure{\label{fig:exp-gpu-model-resnet}\includegraphics[width=0.41\columnwidth]{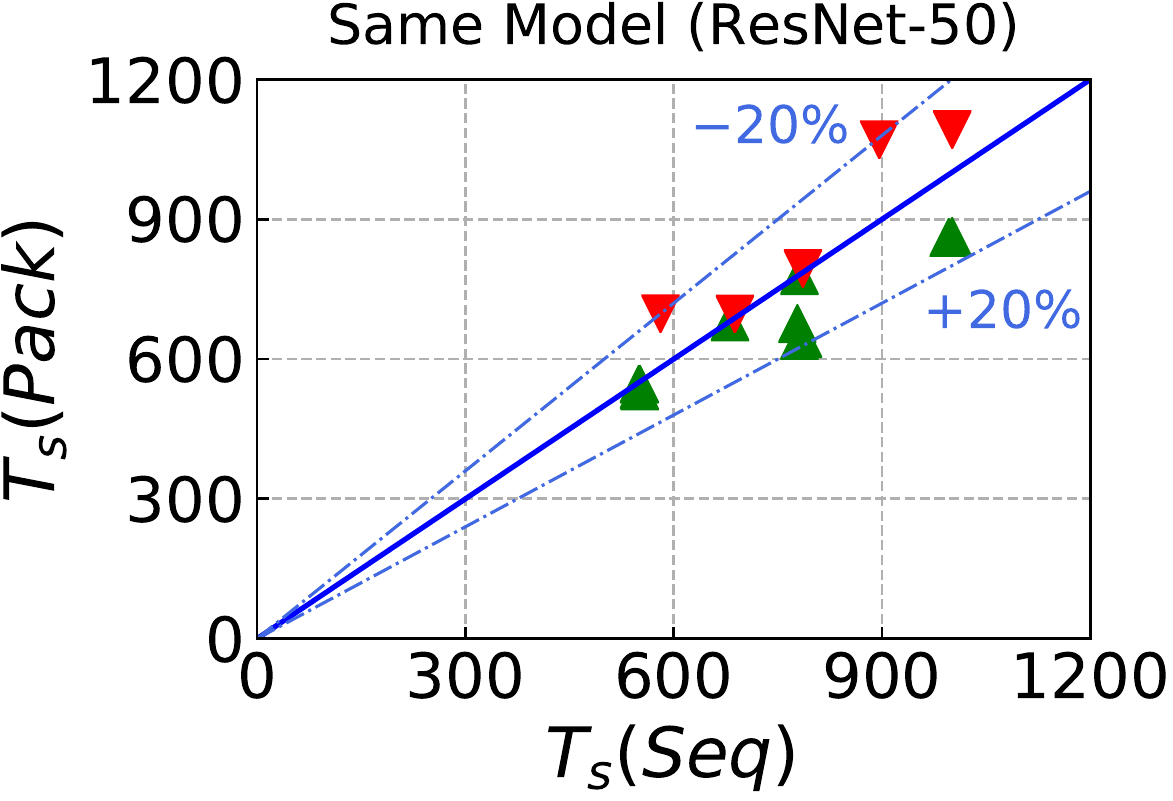}}\hspace{-3pt}\vspace{-1pt}
    \subfigure{\label{fig:exp-gpu-data-resnet}\includegraphics[width=0.41\columnwidth]{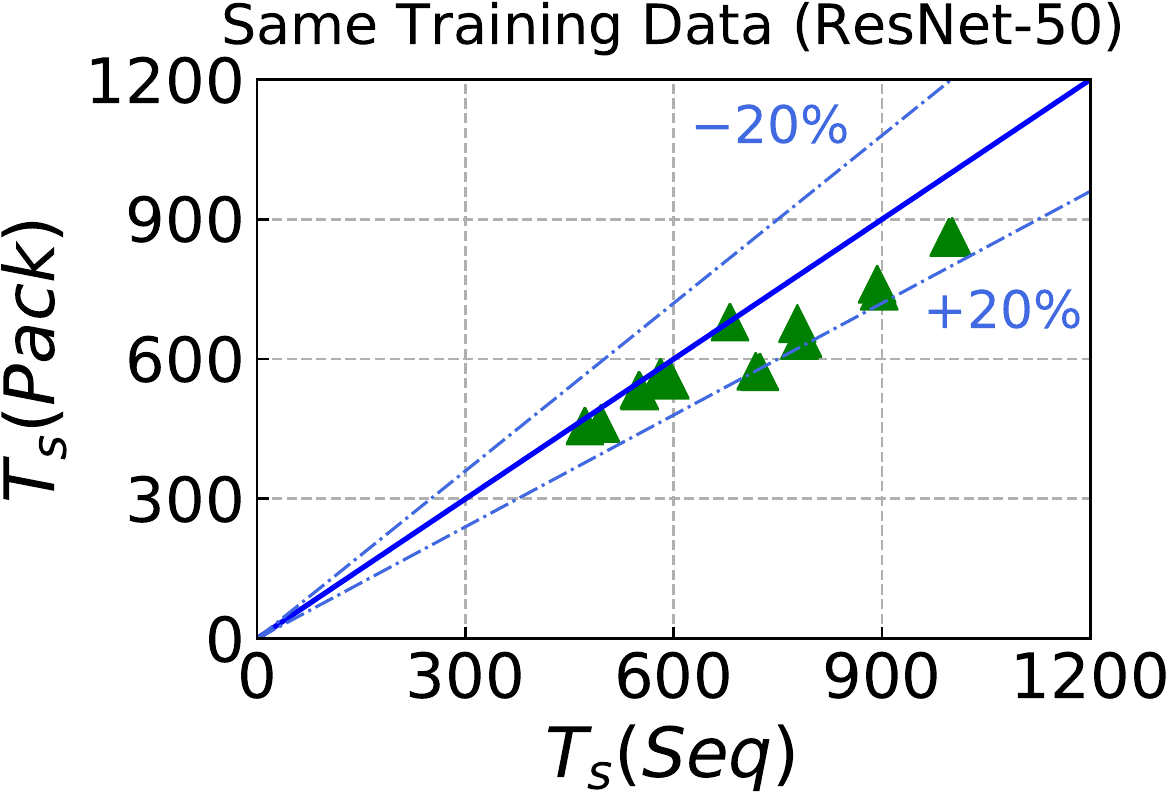}}\hspace{-3pt}\vspace{-1pt}
    \subfigure{\label{fig:exp-gpu-prep-resnet}\includegraphics[width=0.41\columnwidth]{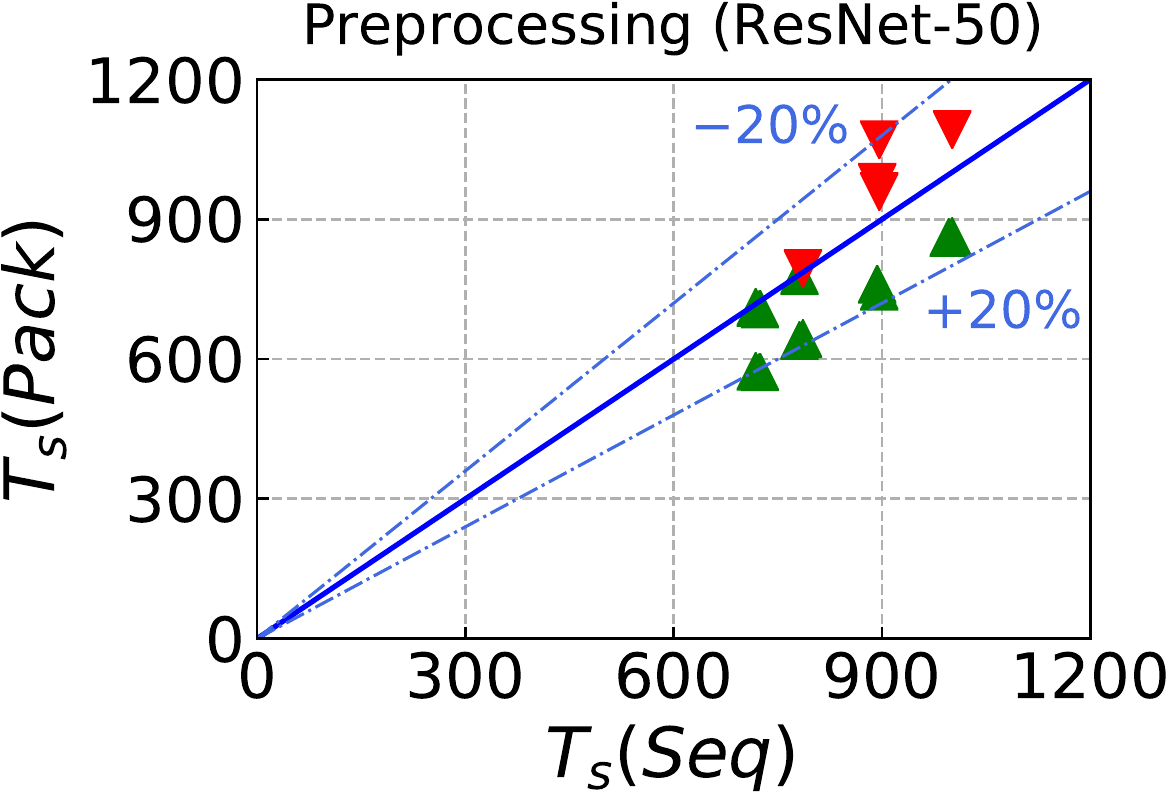}}\hspace{-3pt}\vspace{-1pt}
    \subfigure{\label{fig:exp-gpu-opt-resnet}\includegraphics[width=0.41\columnwidth]{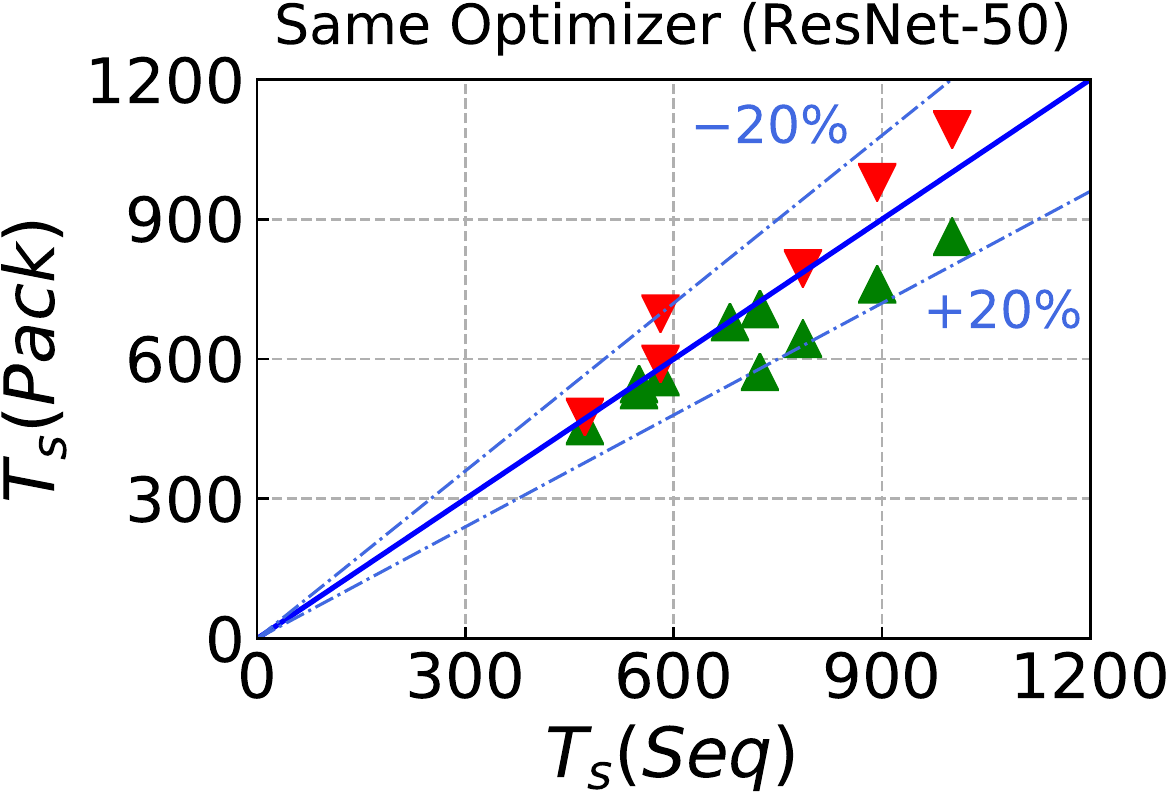}}\hspace{-3pt}\vspace{-1pt}
    \subfigure{\label{fig:exp-gpu-bs-resnet}\includegraphics[width=0.41\columnwidth]{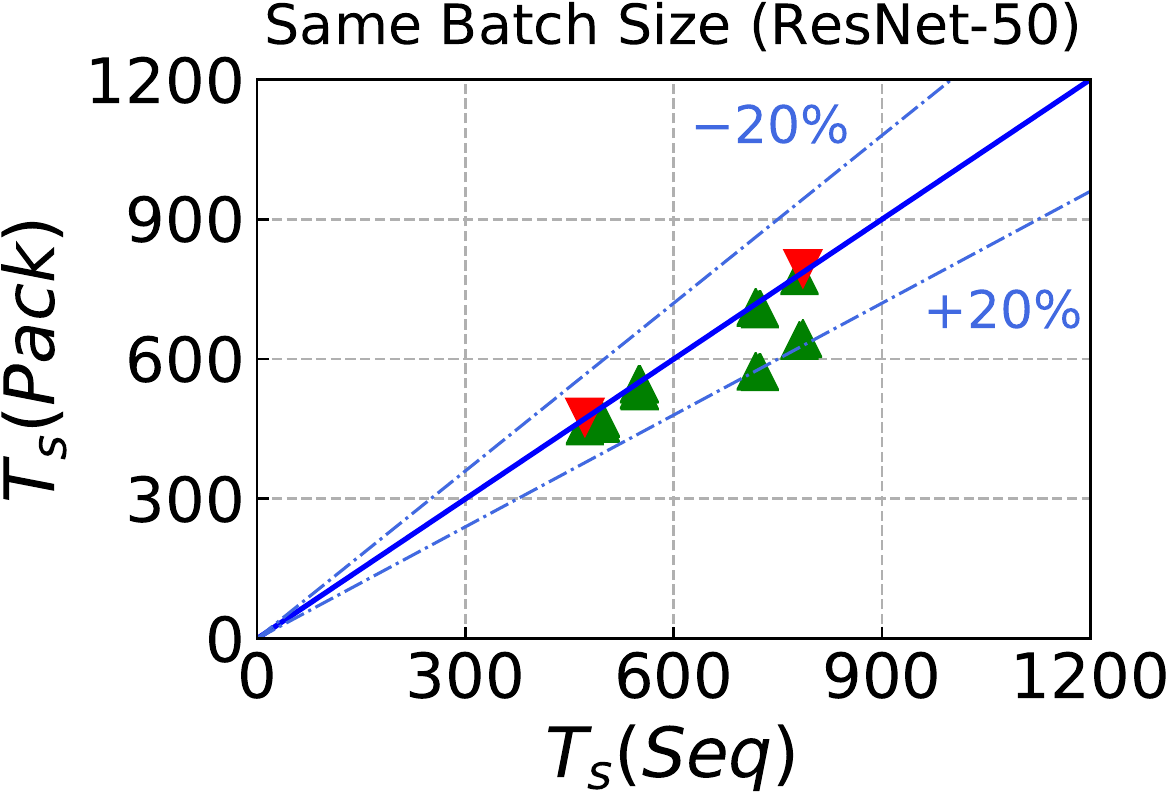}}\hspace{-3pt}\vspace{-1pt}
    \subfigure{\label{fig:exp-gpu-model-densenet}\includegraphics[width=0.41\columnwidth]{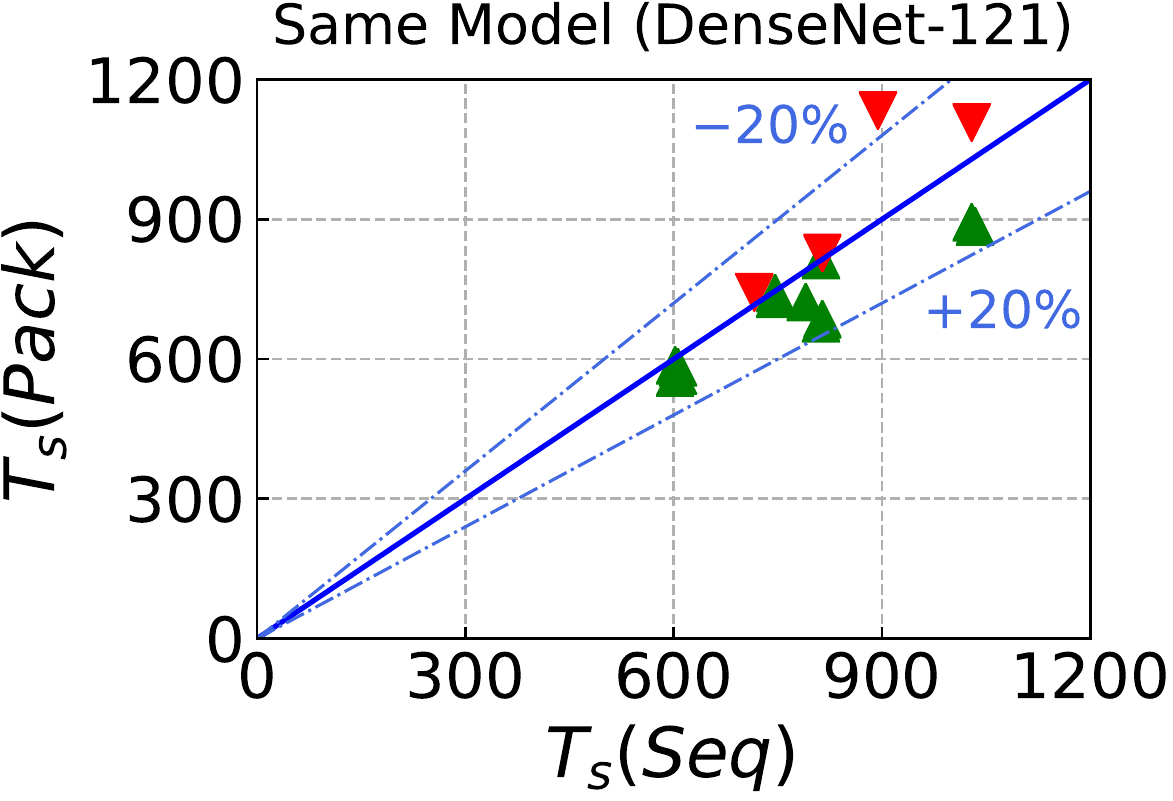}}\hspace{-3pt}
    \subfigure{\label{fig:exp-gpu-data-densenet}\includegraphics[width=0.41\columnwidth]{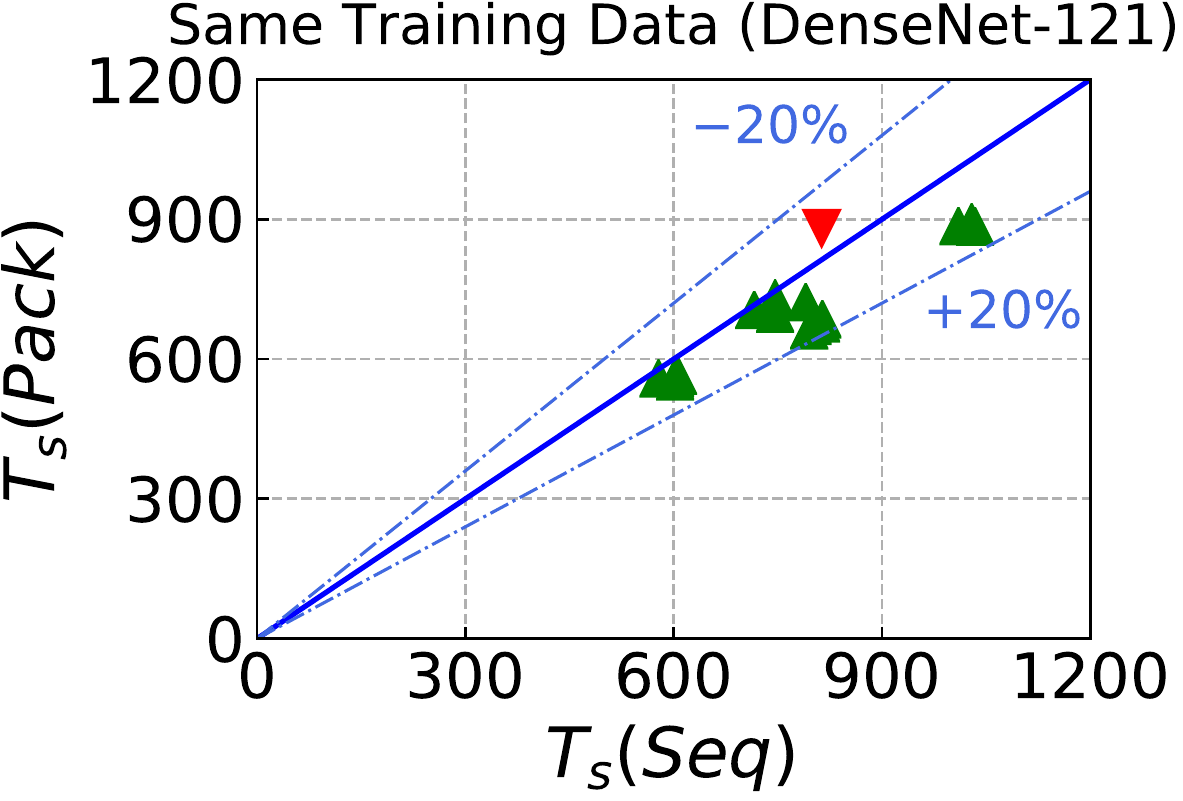}}\hspace{-3pt}
    \subfigure{\label{fig:exp-gpu-prep-densenet}\includegraphics[width=0.41\columnwidth]{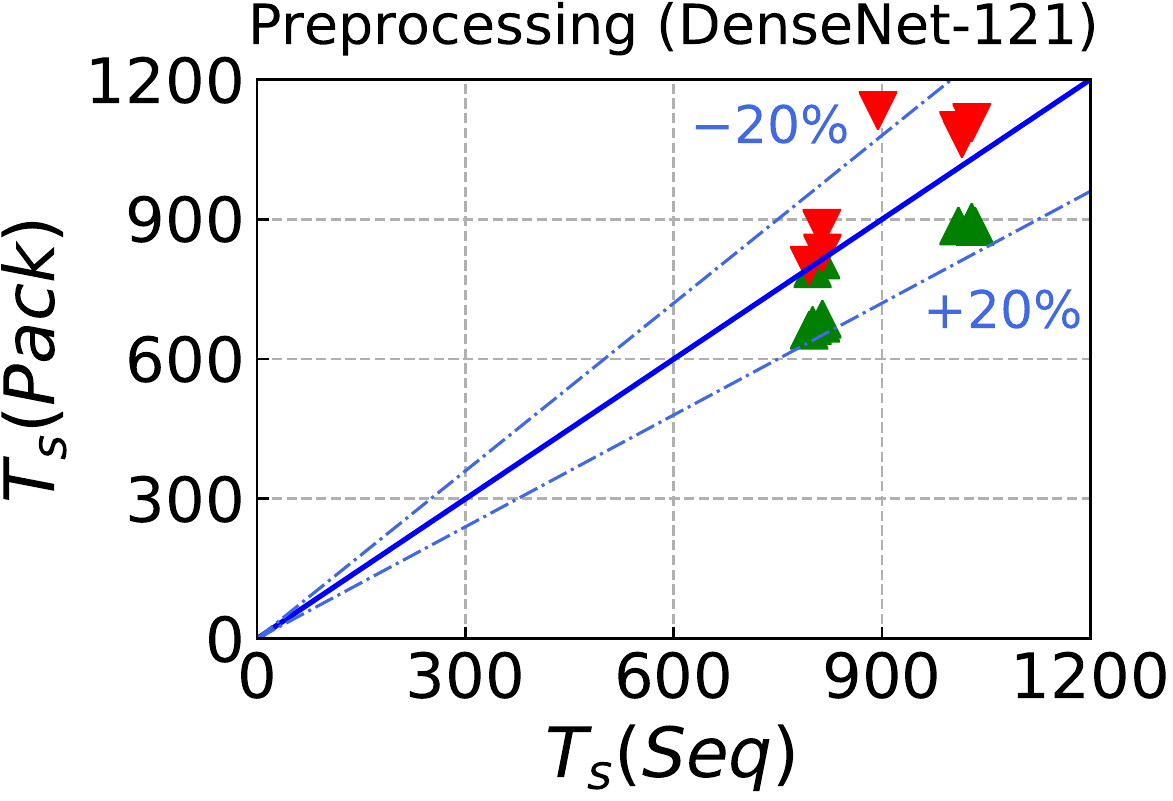}}\hspace{-3pt}
    \subfigure{\label{fig:exp-gpu-opt-densenet}\includegraphics[width=0.41\columnwidth]{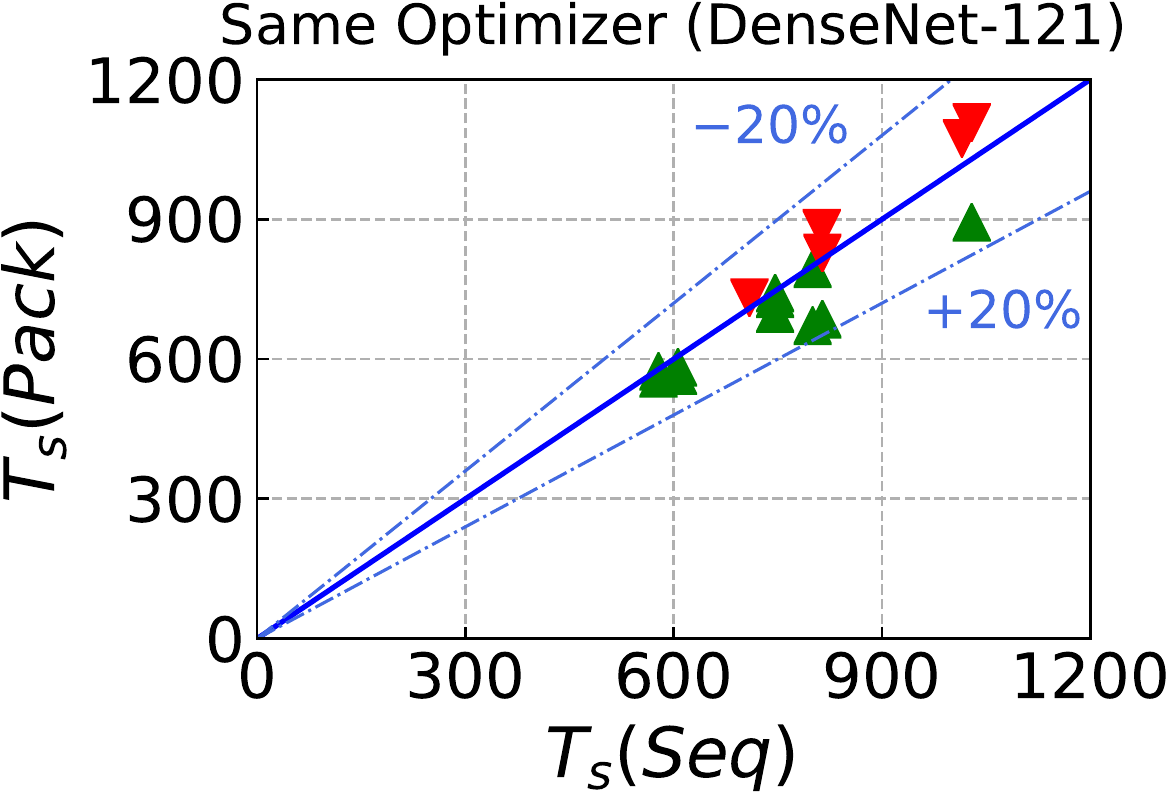}}\hspace{-3pt}
    \subfigure{\label{fig:exp-gpu-bs-densenet}\includegraphics[width=0.41\columnwidth]{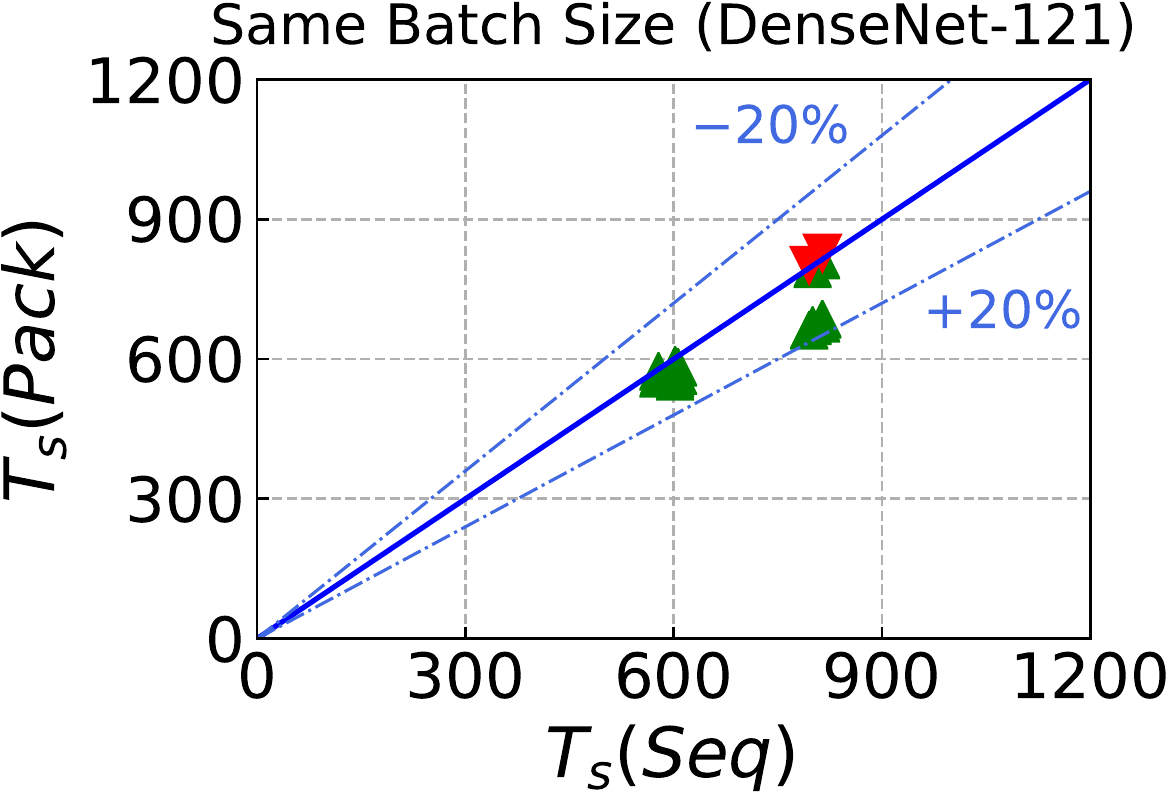}}
    \vspace{-10pt}
    \caption{$T_s(Seq)$ vs. $T_s(Pack)$ (milliseconds) when packing two models on a GPU (NVIDIA Quadro P5000)}
    \label{fig:exp-micro-gpu}
\end{figure*}

\subsection{Ablation Study}
To further evaluate the performance of packing models on GPU, we test more cases based on the five factors: (1) whether the models have the same architecture; (2) whether the models share the same training data; (3) whether the models take the preprocessed data or raw data for training, i.e., if the preprocessing is included in training; (4) whether the models use the same optimizer; and (5) whether the models have the same training batch size. In this ablation study, we follow the configurations illustrated in Table~\ref{tab:micro-conf} and evaluate the \texttt{pack} primitive. Without loss of generality, we focus on packing two models to understand the relationship between the training time and the above factors, packing more models follows the trends as demonstrated in Figure \ref{fig:exp-gpu-improve}.

Figure \ref{fig:exp-micro-gpu} presents the results of the ablation study. In the figure, the data points in each sub-figure represent the $T_s(Seq)$ and $T_s(Pack)$ of various configurations with fixing one configuration (e.g., same batch size or same model). The red point (triangle pointed down) indicates that packing two models brings more overhead compared with training them sequentially with this configuration, i.e., $T_s(Pack) > T_s(Seq)$, while the green point (triangle pointed up) means the opposite. The further from the line, the more significant the performance difference.

As we can see from Figure \ref{fig:exp-micro-gpu}, the best scenarios are where the same training data and the same batch size are used. Over all the configurations, the \texttt{pack} primitive always brings benefits when we train models with the same data since it will reduce the data transfer. Similar benefits happen with the same batch size configuration.
This is important to note because even when the same models are trained but with different data inputs and batch sizes, there can be significant downsides to packing.
It is not simply a matter of looking at the neural network architecture, but the actual training procedure factors into the decision of packing.


\section{\texttt{Pack}-Aware Hyperparameter Tuning}
We demonstrate how \texttt{pack} benefits hyperparameter tuning in this section. As we show in the previous section, \texttt{pack} brings the biggest improvement when the models trained are similar and train on the same input data. Such a scenario naturally arises in hyperparameter tuning. Developers have to search over adjustable parameters such as batch sizes, learning rates, optimizers, etc. Tuning such hyperparameters is crucial to finding models that generalize to unseen data and achieve promising accuracy. 

There are a number of real-world scenarios where multiple models are trained on the same data, we demonstrate hyperparameter tuning as a representative application. Furthermore, \texttt{pack} is a \textit{simple but practical} mechanism that can be implemented at application level, which allows for a wide variety of deployment scenarios.

\subsection{Hyperband}
We explore how we can extend a state-of-the-art hyperparameter tuning algorithm, Hyperband~\cite{DBLP:journals/jmlr/LiJDRT17}, to better share GPU resources. 
Hyperband works by repeatedly sampling random parameter configurations, partially training models with those configurations and discarding those configurations that do not seem promising.
Prior work suggests that Hyperband is effective for parallel hyperparameter search in comparison to sequential algorithms such as Bayesian Optimization~\cite{DBLP:journals/corr/abs-1810-05934}. 

Hyperband poses the search as an online resource allocation problem.
Given $N$ discrete model configurations to test, it partially trains each configuration and discards those that do not seem promising based on a technique called successive halving.
The search routine follows the structure of Algorithm \ref{alg:hyperband}.

\IncMargin{1em}
\begin{algorithm}
\small
\SetKwData{Left}{left}\SetKwData{This}{this}\SetKwData{Up}{up}
\SetKwFunction{Union}{Union}\SetKwFunction{FindCompress}{FindCompress}
\SetKwInOut{Input}{input}\SetKwInOut{Output}{output}
\Input{$R$, $\eta$}
\Output{Conf with the smallest intermediate loss so far}
\For{$r\leftarrow 0$ \KwTo $\lfloor log_{\eta}(R) \rfloor$}{
\emph{Randomly sample $T$ from $N$ confs without replacement}\;
\For{$i\leftarrow 0$ \KwTo $r$}{
Train \textit{conf} $i$  for multiple epochs\;
Calculate the intermediate loss of \textit{confs} $i$\;
Keep a fraction of the best \textit{confs} for the next iteration\;
}
}
\caption{Hyperband}\label{alg:hyperband}
\end{algorithm}\DecMargin{1em}

Intuitively, Hyperband only allocates resources to the most promising configurations. 
At the maximum iteration, the most promising configurations are trained for the longest.
This basic loop can be trivially distributed a random partition of $N$ configurations.
Although Hyperband is able to optimize the process of hyperparameter tuning, the algorithm is long-running since it consists of a large number of trial hyperparameter configurations to run and each of them usually occupies the entire GPU resource when running.

\subsection{\texttt{Pack}-aware Hyperband}
Our \texttt{pack} primitive allows Hyperband to jointly train configurations when possible thereby reducing the overall training time. We propose a \texttt{pack}-aware Hyperband that leverages model packing to improve its performance when there are more models to evaluate than available GPU devices. The challenge is to determine which configurations to train jointly and which to train sequentially.

For each iteration, our \texttt{pack}-aware Hyperband will partition sampled $T$ models to multiple packed groups that can fit on a single device (the size of packed group do not exceed the amount of memory of the GPU). Then, the optimization problem is to search over all packable groups to find the best possible configuration (one that maximizes the overall run time). Note that the singleton partitioning (every single model forms a group) is always a viable solution and potentially even an optimal solution in some cases. We call this primitive \texttt{pack\_opt}, which solves the search problem by producing feasible packing groups and identifying the most promising configuration. Accordingly, we can run a modified Hyperband loop that packs models when beneficial, as shown in Algorithm \ref{alg:hyperband_pack}.

\IncMargin{1em}
\begin{algorithm}
\small
\SetKwData{Left}{left}\SetKwData{This}{this}\SetKwData{Up}{up}
\SetKwFunction{Union}{Union}\SetKwFunction{FindCompress}{FindCompress}
\SetKwInOut{Input}{input}\SetKwInOut{Output}{output}
\Input{$R$, $\eta$}
\Output{Conf with the smallest intermediate loss so far}
\For{$s\leftarrow 0$ \KwTo $\lfloor log_{\eta}(R) \rfloor$}{
\emph{Randomly sample $T$ from $N$ confs without replacement}\;
packed\_group $\leftarrow$ \texttt{pack\_opt(T)}\;
\For{$g\leftarrow 0$ \KwTo packed\_group}{
Train \textit{packed\_conf} $g$ for multiple epochs\;
Calculate the intermediate loss of \textit{packed\_conf} $g$\;
Keep a fraction of the best \textit{confs} for the next iteration\;
}
}
\caption{\texttt{Pack}-aware Hyperband}\label{alg:hyperband_pack}
\end{algorithm}\DecMargin{1em}

There are two challenges in \texttt{pack\_opt}: \textit{(C1)} developing an accurate cost model to evaluate the cost of a packed plan, and \textit{(C2)} a search algorithm that can effectively scale with $T$.
Of course, the combinatorial nature of this problem makes both \textit{(C1)} and \textit{(C2)} hard to accomplish optimally and we need a heuristic to address this problem.
Recognizing that similar models could be packed well together, we design a nearest-neighbor based heuristic.

The method randomly selects a single configuration (out of $T$ for each round) as the centroid and packs all other configurations similar to it until the device runs out of memory. This process is repeated until all models are packed or determined that the best choice is to run them sequentially. 
For calculating the similarity, we map hyperparameter configurations to multi-dimensional feature space and measure the pairwise Euclid distance among all the configurations.
A user-tuned similarity threshold decides how aggressively the system will pack models.
For example, considering the sampled hyperparameter configurations is shown in the Table \ref{tab:hp_conf_exp}, we take standard distance unit as 1, and compute the distance between any two configurations. For categorical hyperparameters like optimizer and activation, the distance is 0 if same and 1 if different, for numeric hyperparameters, we use the index to compute distance.  So, the distance between configuration \textit{A} [batch size:20, optimizer: SGD, learning rate:0.01, activation: ReLu] and configuration \textit{B} [batch size:40, optimizer: Adagrad, learning rate:0.01, activation: ReLu] is $5$. 

Despite being imperfect, Euclid distance has been proven to be a practical metric. We also applied a pairwise Training Time-based distance that reflects the importance of all features using the training time metric. Specifically, we train two configurations in a packed way and a sequential way for a single step respectively, measuring the training time, and calculating the difference with normalization. We take the difference as the distance and deploy it to the pack-ware hyperband. Our empirical experiments show that the Euclid distance method is still faster than Training Time-based distance method up to 18\%.

Note that since the main benefit of \texttt{pack} comes from sharing and padding the input, packing different models can still improve the performance. So, \texttt{pack} is performant in the exploration phase of various hyperparameter tuning methods. Taking Bayesian Optimization as an example, in its exploration phase, the hyperparameter configurations are sampled and evaluated for some predefined objective functions. Thus, the sampled configurations can be packed during the exploration for accelerating.

\subsection{Evaluation for Hyperparameter Tuning}
The goals of our evaluation are two-fold: first to demonstrate that \texttt{pack} can significantly improve hyperparameter tuning performance and second to evaluate our \texttt{pack}-aware Hyperband.
We conduct the experiments based on the same hardware environment as illustrated in section \ref{subsec:setup}. We examine the Hyperband variants on CIFAR-10 \cite{krizhevsky2009learning} which consists of $60000$ color $32 \times 32$ images in 10 classes ($50000$ for training dataset, $10000$ for testing dataset). 
The system's goal is to find the best configuration of those described in Table \ref{tab:hp_conf_exp}, thus all hyperparameter configurations are from the combination of all hyperparameters which has 1056 configurations in total. The input, $R$ and $\eta$, are set to $81$ and $3$, according to the original Hyperband paper \cite{DBLP:journals/jmlr/LiJDRT17}.    

\begin{table}[ht!]
\newcolumntype{?}{!{\vrule width 2pt}}
\onehalfspacing
\small
\centering
\begin{tabular}{c|c}
\Xhline{3\arrayrulewidth}
\hline
\hline
Hyperparameter & Value \\
\hline
Batch size & 20, 25, 30, 35, 40, 45, 50, 55, 60, 65, 70 \\
\hline
Optimizer & Adam, SGD, Adagrad, Momentum \\
\hline
Learning Rate & 0.000001, 0.00001, 0.0001, 0.001, 0.01, 0.1 \\
\hline
Activation & Sigmoid, Leaky ReLu, Tanh, ReLu \\ 
\hline
\hline
\Xhline{3\arrayrulewidth}
\end{tabular}
\caption{Hyperparameter configurations for evaluation}
\vspace{-12pt}
\label{tab:hp_conf_exp}
\end{table}

\noindent We also compare our \texttt{pack}-aware Hyperband against two other heuristics:

\vspace{0.25em} \noindent \textbf{Random Pack Hyperband}: After sampling hyperparameter configurations, the method randomly selects $m$ configurations to pack and evaluates them together, then it keeps the best $n$ configurations and discards the rest as the original Hyperband does. 

\vspace{0.25em} \noindent \textbf{Batch-size Pack Hyperband}: Rather than randomly selecting, \textit{Batch-size Pack Hyperband} only packs the models with the same batch size. Although the number of packed models is confined by GPU memory size, greedy method is employed (i.e. packing as many models as possible until full usage of GPU memory). 

\vspace{0.5em}

We evaluate the overall running time of Hyperband with the different \texttt{pack\_opt} algorithms. 
As presented in the Table \ref{tab:hp_result}, all the \texttt{pack}-aware Hyperband variants can reduce the running time w.r.t the original Hyperband algorithm for all scenarios.   
Our proposal, \textit{kNN Pack Hyperband}, achieves the best performance since it takes advantage of our findings from the previous section where packing the most similar models leads to the biggest improvements. The conclusion is that such an approach can save time (and consequently money) in real end-to-end tasks.
A simpler heuristic, \textit{Batch-size Pack Hyperband}, is not as effective because it under-utilizes the available GPU resources by missing packing opportunities with models with slightly different batch sizes. 
%
To emphasize this point, a \textit{Random Pack Hyperband} can save more time than \textit{Batch-size Pack Hyperband} since it achieves a better GPU resource utilization. 
Our kNN strategy gets the best of both worlds: it finds the most beneficial packing opportunities while completely utilizing the available resources, and benefits are scalable when deployed in an environment with a larger GPU resource.
\begin{table}[htbp]
\newcolumntype{?}{!{\vrule width 2pt}}
\onehalfspacing
\small
\centering
\setlength\tabcolsep{2pt}
\begin{tabular}{P{1.9cm}|c|c|c|c|c}
\Xhline{3\arrayrulewidth}
\hline
\hline
 & Original & Batch-size & Random & kNN & Speedup \\
\hline
MLP-3 & 9236s & 5260s & 3682s & 3491s & $\sim$ \textbf{2.7$\times$} \\
MobileNet & 52092s & 45787s & 36973s & 30182s & $\sim$ \textbf{1.7$\times$} \\
ResNet-50 & 98067s & 89162s & 75436s & 70047s & $\sim$ \textbf{1.4$\times$} \\
DenseNet-121 & 131494s & 126437s & 117405s & 108673s & $\sim$
\textbf{1.2$\times$} \\
\hline
\hline
\Xhline{3\arrayrulewidth}
\end{tabular}
\caption{Performance of \texttt{pack}-based Hyperband}
\vspace{-12pt}
\label{tab:hp_result}
\end{table}



\section{Related Work}

There are number of systems that attempt to control resource usage in machine learning, specifically memory optimization~\citep{narayanan2018accelerating, wang2018superneurons, zhang2019efficient, le2018tflms, jin2018layer, sekiyama2018profile, DBLP:conf/mlsys/YuC20, DBLP:conf/cidr/BoehmADGIKLPR20}, but we see this problem as complementary. For example, \texttt{pack} is similar in mechanism to a recent proposal, HiveMind \citep{narayanan2018accelerating}, where multiple models are fused into a single computational graph during training. However, we additionally contribute: (1) a cost-model and optimizer that decides when this fusion is most beneficial, (2) integration with a hyperparameter tuning algorithm to demonstrate end-to-end improvements over a training workload, and (3) a data batching scheme that allows packing models with different batch sizes without hurting statistical efficiency. These contributions are noted as existing limitations in HiveMind.

We also discuss the related works that study hyperparameter tuning systems and multi-tenancy systems in machine learning.

\subsection{Systems for Hyperparameter Tuning}

Since hyperparameter tuning is a crucial part of the machine learning development process, a number of systems have been proposed to scale up such search routines. 
For example, Google Vizier~\citep{golovin2017google} exposes hyperparameter searching as a service to its organization's data scientists. Aggressive "scale-out" has been the main design principle of Vizier and similar systems~\citep{liaw2018tune, li2018massively, DBLP:journals/jmlr/LiJDRT17}.

Recently, there has been a trend toward more controlled resource usage during hyperparameter tuning. Cerebro borrows the idea of multi-query optimization in database system to raise resource efficiency \cite{DBLP:journals/pvldb/NakandalaZK20}. HyperSched proposes a scheduling framework for hyperparameter tuning tasks when there are contended specialized resources \cite{liaw2019hypersched}.
And, some work has been done on resource management~\citep{sparks2015automating} and pipeline re-use~\citep{li2019exploiting} in the non-deep learning setting.
We believe that \texttt{pack} and \texttt{pack\_opt} are two primitives that are useful in hyperparameter tuning when specialized hardware such as GPUs and TPUs are limited in usage. Also, although our experiments focus on hyperparameter tuning, \texttt{pack} and \texttt{pack\_opt} primitives can be easily extended to other scenarios. 

\subsection{Systems for Multi-tenancy}

Most current projects about building multi-tenant systems for machine learning deployment is based on device-level placement, i.e., dividing resources at the granularity of full devices (e.g., an entire server or GPU). Here, the scheduler partitions a cluster of servers where each server has one or more GPUs for various model training tasks and seeks to reduce the overall training time by intelligent placement. Other scheduling methods have followed, such as Tiresias~\citep{gu2019tiresias} and Optimus~\citep{peng2018optimus}. Several extensions have been proposed to this basic line of work including fairness~\citep{mahajan2019themis}, preemption~\citep{yabuuchi2019low}, and performance prediction~\citep{zheng2019cynthia}. Gandiva is a cluster scheduling framework for deep learning jobs that provides primitives such as time-slicing and migration to schedule different jobs. CROSSBOW is a system that enables users to select a small batch and scale to multiple GPUs for training deep learning models \citep{DBLP:journals/pvldb/KoliousisWWMCP19}. PipeDream is a deep neural network training system for GPUs that parallelizes computation by pipelining execution across multiple machines that partitions and pipelines training jobs across worker machines \citep{narayanan2019pipedream}. Ease.ml is a declarative machine learning service platform that focuses on a cost-aware model selection problem in a multi-tenant system. \citep{DBLP:journals/pvldb/LiZLWZ18}. Some recent works also exploit data parallelism to accelerate the training process. MotherNets can ensembles different models and accelerate the training process by reducing the number of epochs needed to train an ensemble \cite{DBLP:conf/mlsys/WasayHLCI20}. FLEET theoretically proves that optimal resource allocation in deep learning training is NP-hard and propose a greedy algorithm
to allocate resources with data sharing \cite{DBLP:conf/mlsys/GuanMSLP20}.

Compared with these previous works, our prototype implements a method that can pack diverse models with different batch sizes. We also conduct a comprehensive evaluation that differentiates performance wins from variable elimination v.s. improved utilization, and highlight potential for packed models to train slower than the sum of their parts, which is only apparent with modern architectures. Take these inspirations, we further deploy our primitive to hyperparameter tuning and present that \texttt{pack} can significantly accelerate it.

\section{Discussion}
\label{sec:dis}

Our core contribution is demonstrating the potential benefits (and overheads) of combining similar models into a single computational graph, and thus collapsing common data inputs during training iterations. This was the reason why we chose not to optimize this process at a lower level (e.g., MPS/Hyper-Q \cite{hyperq}), where we found that the majority of benefits could be attributed to simply sharing common inputs and context variables. Thus, the key goal of our proposed optimizer is to decide whether two models share enough to see a potential benefit, and controlling the exact execution order of the computation is orthogonal to our contribution.

Our long-term goal is to build a system for multi-tenant deep learning deployment, and we believe the \texttt{pack} will be one of the core parts of multi-tenancy systems for machine learning. In hyperparameter tuning there is a single-user and a clear SLO (find the best model configuration over all), then to extend to more general multi-tenancy settings where concurrent models are trained, we will reason about multiple users, priorities, and user-specified objectives. For this, we decide to first make a deep investigation on a single GPU so that we will know how to optimize when there are multiple GPUs. Thus, any distributed training and the regarding optimization is out of the scope of the paper.

We implemented \texttt{pack} on TensorFlow to conduct a comprehensive evaluation and highlight its benefits in hyperparameter tuning. Although we believe that a custom execution platform could improve performance, \texttt{pack} doesn't require the modification of any specific framework and can be implemented across frameworks. We focus \texttt{pack} as a higher-level primitive due to (1) the optimizations will be more transferable across ML execution frameworks and thus increase the impact or applicability of our insights, and (2) many low-level libraries are highly optimized and introducing these changes (e.g. supporting jagged arrays) we believe are interesting research questions on their own.




\section{Conclusion and Future Work}
We analyze the benefits and limitations of packing multiple models together to take advantage of available GPU resources for model training. Under the proper conditions, this packing can bring up to 40\% reduction in latency per model packed, compared with training the models sequentially on a GPU. 
We further demonstrate that \texttt{pack} primitive can be used to accelerate a state-of-the-art hyperparameter tuning algorithm.
Our end-to-end tuning system demonstrates a $2.7$x speedup in terms of time to find the best model by improving GPU utilization.
Our analysis opens many interesting optimization opportunities, such as the training process can be decomposed and scheduled for packing to reduce the overall training time, or trading off accuracy or training time to improve overall resource utilization. 


\bibliographystyle{ACM-Reference-Format}
\bibliography{ref}


\end{document}